\documentclass[10pt,twocolumn,letterpaper]{article}

\usepackage[dvipsnames, table,xcdraw]{xcolor}

\usepackage[pagenumbers]{cvpr} 

\newcommand{\model}{\textsc{Robin}\xspace}
\newcommand{\dataset}{\textsc{SVG}\xspace}
\newcommand{\distill}{\textsc{SG-Edit}\xspace}
\newcommand{\datasetfirst}{\textsc{SVG-Relations}\xspace}
\newcommand{\datasetsecond}{\textsc{SVG-SG}\xspace}

\newcommand{\modelname}{Robin\xspace}

\usepackage{etoolbox, array}
\usepackage[font=footnotesize,labelfont=bf,aboveskip=3pt,belowskip=-8pt]{caption}

\newdimen\abovecrulesep
\newdimen\belowcrulesep
\makeatletter
\patchcmd{\@@@cmidrule}{\aboverulesep}{\abovecrulesep}{}{}
\patchcmd{\@xcmidrule}{\belowrulesep}{\belowcrulesep}{}{}

\definecolor{demphcolor}{RGB}{144, 144, 144}
\definecolor{mygray}{gray}{0.4}
\definecolor{lightgray}{rgb}{0.9, 0.9, 0.9}

\newlength\savewidth

\newcommand{\tablestyle}[2]{\setlength{\tabcolsep}{#1}\renewcommand{\arraystretch}{#2}\centering\footnotesize}
\makeatletter\renewcommand\paragraph{\@startsection{paragraph}{4}{\z@}{.5em\@plus1ex\@minus.2ex}{-.5em}{\normalfont\normalsize\bfseries}}
\makeatother

\definecolor{cvprblue}{rgb}{0.21,0.49,0.74}
\usepackage[pagebackref,breaklinks,colorlinks,allcolors=cvprblue]{hyperref}
\usepackage{graphicx} 
\usepackage{microtype}
\usepackage{url}
\usepackage{ulem} 
\usepackage{tcolorbox}
\usepackage{listings}
\usepackage{xspace}
\usepackage{multirow}
\usepackage{lipsum}

\usepackage{booktabs}
\usepackage{subcaption}
\usepackage{tabularx}

\usepackage{amssymb}
\usepackage{pifont}
\usepackage{float}
\usepackage{rotating}

\definecolor{LightGray}{gray}{0.9}

\definecolor{pastelblue}{RGB}{173,216,230}
\definecolor{pastelpink}{RGB}{255,182,193}
\definecolor{pastelgreen}{RGB}{152,251,152}

\newtcolorbox{pastelbox}[2][]{colback=#1!10!white, colframe=#1!80!black, 
    boxrule=0.5mm, 
    arc=1.5mm, 
    auto outer arc,
    left=1mm,
    right=1mm,
    top=1mm,
    bottom=1mm,
    title=#2}
    
\usepackage{wrapfig}
\usepackage{amsmath}

\def\paperID{7540} 
\def\confName{CVPR}
\def\confYear{2025}

\title{Synthetic Visual Genome}

\author{
\vspace{1mm}
\textbf{Jae Sung Park}\textsuperscript{1},
\textbf{Zixian Ma}\textsuperscript{1},
\textbf{Linjie Li}\textsuperscript{1},
\textbf{Chenhao Zheng}\textsuperscript{1},
\textbf{Cheng-Yu Hsieh}\textsuperscript{1}, \\
\vspace{1mm}
\textbf{Ximing Lu}\textsuperscript{1},
\textbf{Khyathi Chandu}\textsuperscript{2},
\textbf{Quan Kong}\textsuperscript{4}, \\
\textbf{Norimasa Kobori}\textsuperscript{4},
\textbf{Ali Farhadi}\textsuperscript{1,2},
\textbf{Yejin Choi}\textsuperscript{3},
\textbf{Ranjay Krishna}\textsuperscript{1,2}
\\\\ 
\textsuperscript{1}University of Washington,
\textsuperscript{2}Allen Institute for Artificial Intelligence, \\
\textsuperscript{3}Stanford University,
\textsuperscript{4}Woven by Toyota
}

\begin{document}

\maketitle
\begin{abstract}
Reasoning over visual relationships—spatial, functional, interactional, social, etc.—is considered to be a fundamental component of human cognition. 
Yet, despite the major advances in visual comprehension in multimodal language models (MLMs), precise reasoning over relationships and their generations remains a challenge. 
We introduce \model: an MLM instruction-tuned with densely annotated relationships capable of constructing high-quality dense scene graphs at scale. 
To train \model, we curate \dataset\footnote{\textbf{\dataset}: Synthetic Visual Genome}, a synthetic scene graph dataset by completing the missing relations of selected objects in existing scene graphs using a teacher MLM and a carefully designed filtering process to ensure high-quality.  
To generate more accurate and rich scene graphs at scale for any image,  we introduce \distill: a self-distillation framework where GPT-4o further refines \model's predicted scene graphs by removing unlikely relations and/or suggesting relevant ones. 
In total, our dataset contains $146K$ images and $5.6M$ relationships for $2.6M$ objects.
Results show that our \model-3B model, despite being trained on less than $3$ million instances, outperforms similar-size models trained on over $300$ million instances on relationship understanding benchmarks, and even surpasses larger models up to 13B parameters. 
Notably, it achieves state-of-the-art performance in referring expression comprehension with a score of $88.9$, surpassing the previous best of $87.4$. 
Our results suggest that training on the refined scene graph data is crucial to maintaining high performance across diverse visual reasoning tasks\footnote{The SVG data, model checkpoints, and code are available at \url{https://synthetic-visual-genome.github.io/}}.

\end{abstract}

\begin{figure}[htb!]
    \centering
    \begin{tabular}{cc}
    \includegraphics[width=0.95\columnwidth]{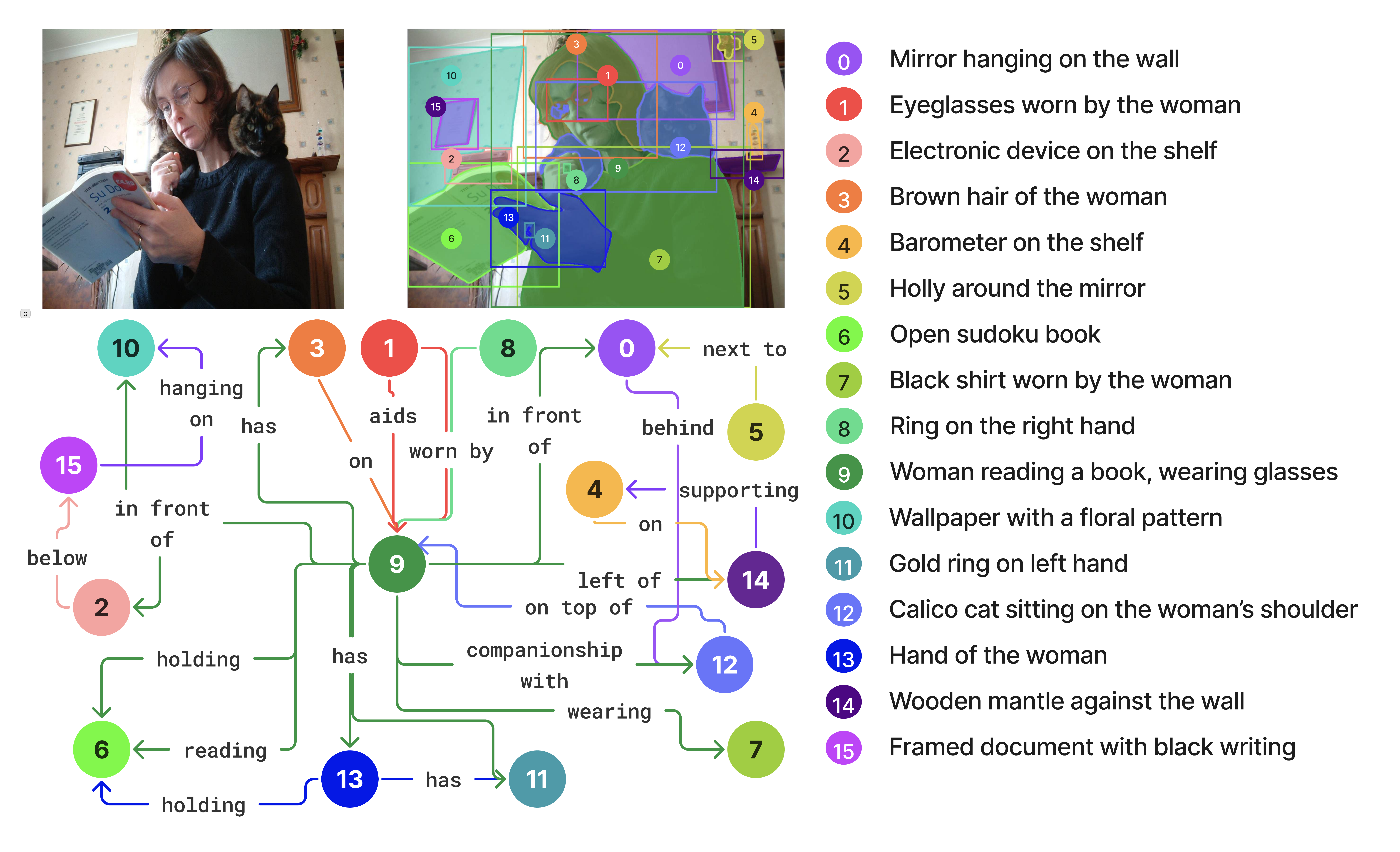}
    
    
    \end{tabular}
    \caption{
    \small
     Example of our Synthetic Visual Genome (SVG) dataset -- the first automatically generated large-scale scene graph dataset with diverse open-set categories, fine-grained regions, and densely annotated relationships. SVG averages four times more relations per object than Visual Genome~\cite{krishna2017visual}.
     }
    \label{fig:teaser}
\end{figure}

\section{Introduction}

\begin{figure*}[t!]
    \centering
    \includegraphics[width=0.8\textwidth]{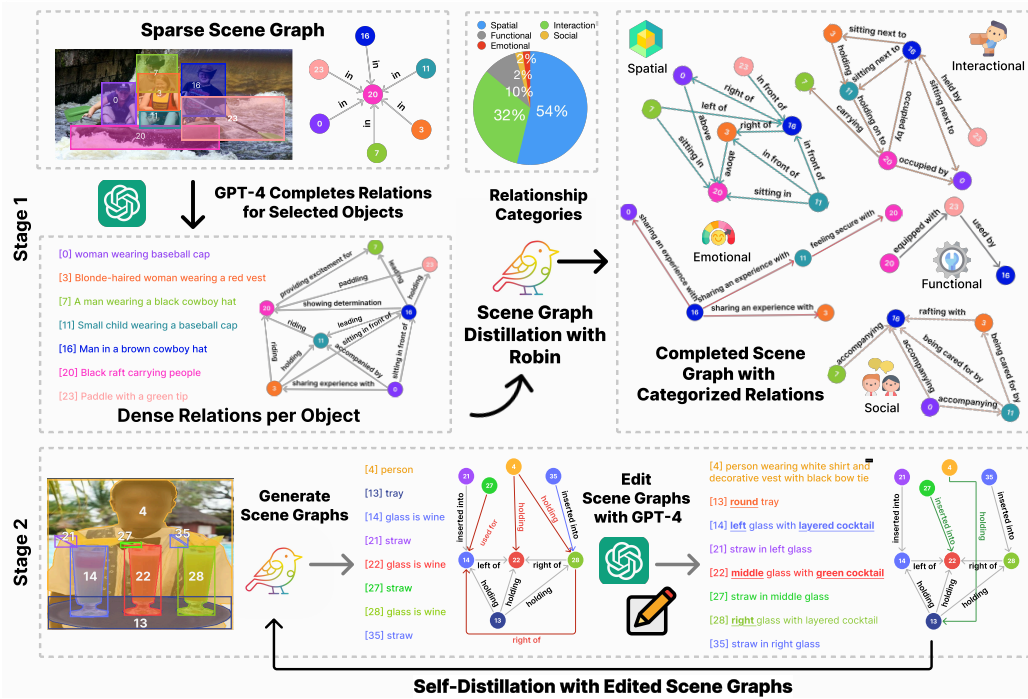}
    \caption{
    \small
     Overview of our Synthetic Visual Genome data engine. \textbf{Stage 1}: GPT-4 completes missing relationships for selected objects in existing image annotations, spanning five categories: spatial, interactional, emotional, functional, and social. \textbf{Stage 2}: Self-distillation pipeline where \model generates dense scene graphs from images with segmentations, and GPT-4 refines these graphs by enriching content and correcting errors (\textit{e.g., red to green edges}) before model training.
     }
    \label{fig:robin_approach}
\end{figure*}

Scholars have argued for decades that \emph{visual relationship reasoning} is a fundamental characteristic of human intelligence~\cite{biederman1987recognition,biederman1982scene}.
By reasoning over relationships, people can make sense of new scenes by stitching together individual objects and their pairwise relationships~\cite{janssen1997compositionality,hupkes2020compositionality,bottou2014machine,chomsky1965some}.
For instance, relationships allow people to describe the photo in Figure~\ref{fig:robin_approach} using a set of spatial (\textit{woman-above-raft}), social (\textit{child-cared\_by-woman}), functional (\textit{man-using-paddle}), interactional (\textit{woman-holding-baby}) and emotional (\textit{man-sharing\_an\_experience\_with-family}) relationships.
For a while now, multimodal language model (MLM) research has sought to develop models that can similarly comprehend such relationships~\cite{krishna2017visual,ji2020action,lu2016visual,GrundeMcLaughlin2021AGQA}. Yet despite all the progress, frontier MLMs still struggle to accurately express relationships; open-source MLMs perform even worse~\cite{wang2024picture}.

Instruction-tuning has proven effective for distilling specific reasoning capabilities in MLMs~\cite{dai2024instructblip}.
However, instruction tuning an open-sourced MLM to understand diverse relationships remains infeasible due to the absence of large-scale relationship-focused datasets; existing efforts are limited to spatial reasoning derived from synthetic data~\cite{chen2024spatialvlm}. Scene graphs could serve as potential sources for instruction tuning because they provide direct annotations of objects and their relationships within scenes. Nevertheless, existing scene graph datasets have limitations in image coverage and relationship diversity.  Visual Genome~\cite{krishna2017visual}, for instance, contains a large number of spatial relationships but is limited in interactional, emotional, or functional ones (Figure~\ref{fig:robin_approach}) and covers merely 100K images. Additionally, its annotations are rather \textit{sparse}, with only 1.5 relations annotated on average for every subject, and fail to enumerate all possible relationships in the scene, including some spatial ones like \textit{woman, in front of, baby}. This is because even for humans, exhaustively annotating relationships for all objects is cumbersome, and scaling such detailed labeling is impractical due to high time and cost demands.

To overcome this bottleneck, we propose automating the process of generating densely annotated scene graphs. One option is to prompt frontier models like GPT-4V to generate instruction-tuning data, which has shown promise for many language and reasoning tasks~\cite{wang2022super}. However, such methods have found limited utility in computer vision as even current frontier models struggle to understand 3D structure and interactions between objects~\cite{fu2024blink}. As a result, when prompted to generate scene graph data from scratch, they hallucinate and produce low-quality data~\cite{wang2024allseeing}.

In this work, we introduce a more reliable dataset bootstrapping framework that enables building high-quality scene graph datasets at scale using richly annotated seed images and frontier models. In particular, 1) we start out by leveraging a frontier model, GPT-4V, to enrich existing scene graph datasets by completing the missing relationships based on the initial human annotations available instead of completely from scratch. This leads to a synthetic yet high-quality and dense scene graph dataset \textbf{\dataset}, with $146$K images annotated with $5.6$M relationships for $2.6$M objects.
2) Using \dataset, we then train \textbf{\model-3B}\footnote{ROBIN: Relation-Object Instruction Tuned Model}, a dedicated MLM to generate dense scene graphs with enhanced relationship understanding and grounded reasoning.
3) Finally, we show how \model can facilitate a scalable synthetic scene graph data generation pipeline, \textbf{\distill}, which uses \model to efficiently generate dense scene graph data from scratch coupled with GPT-4o for further refinement (Figure~\ref{fig:robin_approach}, bottom).

Our experiments validate our contributions in three aspects. First, we confirm the quality of \dataset by showing that training with \dataset produces an effective MLM capable of generating high-quality dense scene graphs.
Second, we show that \distill provides a scalable solution for generating large-scale scene graph datasets. Specifically, the resulting enhanced \model-3B, despite being trained on fewer than 3M instances, outperforms similarly sized models trained on over 300M instances and even larger models (up to 13B parameters) across various relationship understanding and grounded reasoning benchmarks.
Finally, we demonstrate that \model-3B achieves state-of-the-art results in open-ended, dense scene graph generation tasks.

\section{Synthetic Visual Genome Data Pipeline}\label{sec:dataset}
Existing scene-graph datasets typically lack dense and diverse relationship annotations, limiting their utility for tasks requiring comprehensive relationship understanding. To address this limitation, we leverage powerful proprietary multimodal models (e.g., GPT-4) to systematically infer missing object relationships. In Figure~\ref{fig:robin_approach}, we propose a two-stage pipeline that gradually enriches and refines the scene graph annotations, which we describe below.

\paragraph{Stage 1: Dense Relationship Completion}\label{sec:dataset_stage1}

In Stage 1, we construct an initial set of densely annotated scene graphs using a carefully selected subset of seed images with reliable, high-quality annotations. While proprietary multimodal models (e.g., GPT-4) are powerful, directly generating scene graphs from scratch can result in hallucinations and grounding errors, making them unsuitable on their own for accurate dataset creation (see Figure~\ref{fig:qual_gpt_sg} in Appendix~\ref{sec:som_scene_graph}). Instead, we try building upon existing human-annotated data to generate scene graphs to ensure correctness. 

Our complete data generation pipeline is illustrated in Figure~\ref{fig:data-gen} (Appendix~\ref{sec:svg_relations_appendix}). We begin with a subset of COCO~\cite{coco} images  that include: a) object detection labels from COCO~\cite{coco} and LVIS~\cite{gupta2019lvis}, b) region descriptions from RefCOCO~\cite{yu2016modeling} and Visual Genome (VG)~\cite{krishna2017visual}, c) scene graphs from VG and GQA~\cite{hudson2019gqa}, and d) depth maps generated by the Depth-Anything model~\cite{depthanything}. This yields 33K seed images with comprehensive annotations for each region. We then select visually meaningful annotated regions by comparing them against masks produced by Segment Anything (SAM)~\cite{kirillov2023segment} and Semantic-SAM~\cite{li2023semantic}, retaining regions with Intersection-over-Union (IoU) scores greater than 0.5. Next, we prompt GPT-4V to generate comprehensive, categorized relationship annotations (spatial, interactional, functional, social, and emotional) for selected prominent objects. Finally, we apply rule-based filtering for spatial relationships and model-based VQA filtering for other relationship types to remove irrelevant or incorrect annotations~\cite{fang2023data, Park2023LocalizedSK}. We refer to this curated dataset as \textbf{\datasetfirst}. Complete implementation details are provided in Appendix~\ref{sec:svg_relations_appendix}.


\paragraph{Stage 2: Dense Scene Graphs with GPT-4 Edits}
\label{sec:dataset_stage2}

While the first stage provides high-quality annotations derived from COCO-based images, its scope of images is limited, and training solely on this data can potentially produce noisy scene graphs for images \textit{in the wild}. Inspired by iterative data improvement methods such as Segment Anything~\cite{kirillov2023segment}, we introduce a distillation-based refinement approach called \textbf{\distill}. We first train a dedicated student model, \textbf{\model}, on \datasetfirst to generate dense relationships for arbitrary images. Directly training on these automatically generated scene graphs though typically requires further intervention to ensure high-quality annotations, a process traditionally handled by humans in~\cite{kirillov2023segment}. Rather than relying on costly human annotators, we leverage GPT-4o as an automated editor to remove noisy relationships, add relevant ones, and enrich object descriptions with precise attributes. 
The bottom part of Figure~\ref{fig:robin_approach} illustrates this editing process, where GPT-4o refines a scene graph by removing incorrect relationships (highlighted in red), adding new relevant relationships (highlighted in green), and specifying detailed attributes for objects (bolded and underlined).


We apply this distillation approach to multiple additional datasets—including ADE20K~\cite{zhou2019semantic}, PSG~\cite{yang2022panoptic}, and VG~\cite{krishna2017visual}—to increase data diversity. Specifically, we use the trained \model~to generate scene graphs from images and segmentation masks in these datasets and then refine them using GPT-4o editing. This process yields a more diverse and extensive synthetic dataset, \textbf{\datasetsecond}, comprising a total of 113K annotated images (25K from ADE20K, 35K from PSG, and 53K from VG).\footnote{Note that while we generated 113K scene graphs in this work, the presented pipeline is scalable; analysis of further scaling is left for future research.}

\begin{table}[ht!]
\centering
\resizebox{\linewidth}{!}{%
\begin{tabular}{llccccc}
\toprule

\multirow{2}{*}{Dataset} & \multirow{2}{*}{Images} & \multirow{2}{*}{Annotator} & \multirow{2}{*}{Region} & Objects &  Triplets & Predicates \\
 & & & & per image & per image & per region \\
\midrule
VG~\cite{krishna2017visual} & 108K & Human & Box & 35.2 & 21.4 & 0.6 \\
GQA~\cite{hudson2019gqa} & 85K & Human & Box & 16.4 & 50.6 & 3.1 \\
OpenImages~\cite{Kuznetsova_2020} & 568K & Human & Box & 8.4 & 5.6 & 0.7 \\
PSG~\cite{yang2022panoptic} & 49K & Human & Seg + Box & 11.2 & 5.7 & 0.6 \\ 
\midrule
\datasetfirst & 33K & GPT-4V & Seg + Box & 13.2 & 25.5 & 1.9 \\
\datasetsecond & 113K & \model + GPT-4o & Seg + Box & 19.8 & 42.3 & 2.4 \\
\bottomrule
                
\end{tabular}}
\caption {
Comparison of our Synthetic Visual Genome datasets (\datasetfirst, \datasetsecond) with existing scene graph benchmarks.
}
\label{tab:dataset_stats}
\end{table}

\paragraph{Comparison with existing datasets.} Table~\ref{tab:dataset_stats} summarizes statistics for both introduced datasets (\datasetfirst, \datasetsecond) in comparison to existing scene-graph benchmarks, highlighting the increased density and diversity of our synthetic annotations.

\section{\model-3B}


Using the \dataset dataset, we introduce \model-3B, an MLM trained to reason accurately about regions and produce dense scene graphs for any objects of interest. Unlike previous MLMs that refer to regions primarily through textual coordinates (e.g., bounding boxes), \model represents each region using both pixel-level segmentation masks~\cite{yuan2024osprey} and their text references~\cite{chen2023shikra, wang2024allseeing, ma2025groma}. This dual representation enables more precise, fine-grained localization of referenced objects. Below, we describe our model's design in detail.

\subsection{Model architecture}  
\begin{figure}[ht!]
    \centering
    \includegraphics[width=0.47\textwidth]{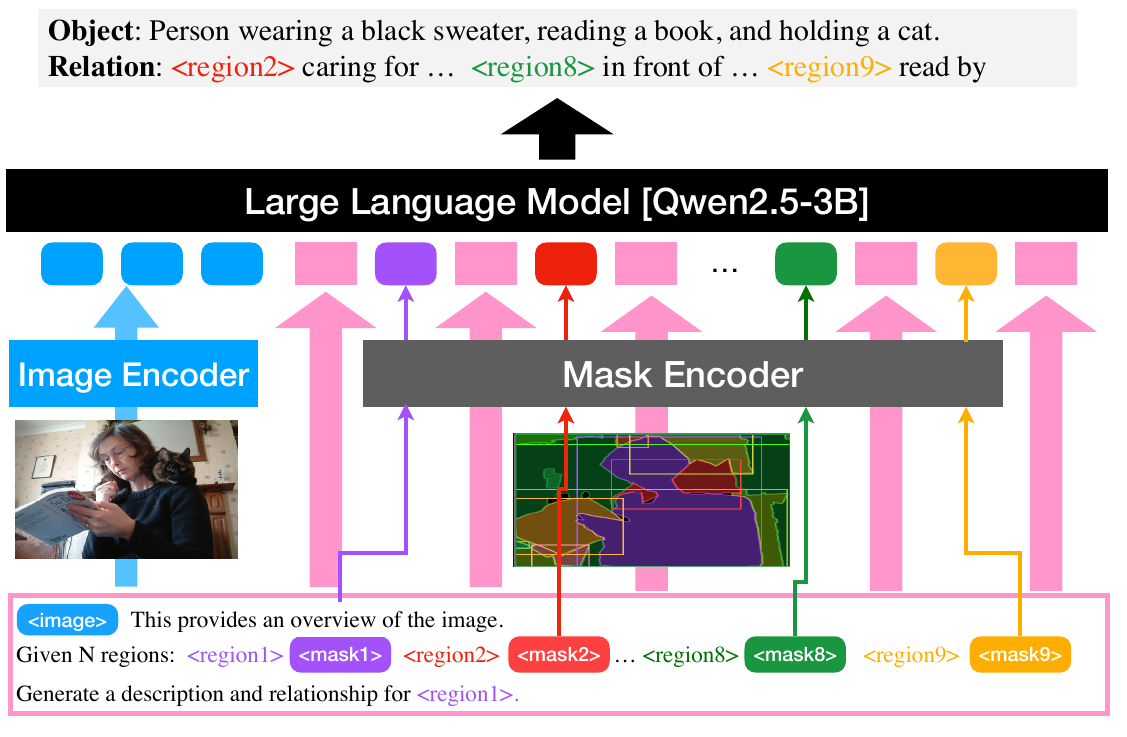}
    \caption{Architecture of \model, consisting of a visual encoder that embeds the global image, a mask encoder that embeds segmentation masks individually, and a text encoder that encodes the textual input. } 
    \label{fig:model_architecture}
\end{figure}

 As shown in Figure~\ref{fig:model_architecture}, our model consists of three components: 1) a vision encoder that encodes the entire image to image tokens, 2) a pixel-level mask-aware extractor that embeds each segmentation to mask tokens, and 3) a language model (LM) that takes in image, mask, and text tokens to support any visual instruction and grounding tasks in the text space, similar to the Osprey model~\cite{yuan2024osprey}. We use ConvNext-Large~\cite{ilharco_gabriel_2021_5143773} as our visual and mask encoder, and Qwen2.5-3B as our LM~\cite{qwen2}. Vision and mask projection layers are added to project the vision and mask tokens into text embeddings that will be passed to the LM. Note that we do not initialize from vision-language model checkpoints and learn their alignment \textit{from scratch}. Using the long context capabilities of our LM (up to 8192 tokens), our model effectively handles up to 99 regions per image.  

\subsection{Training stages}
\label{sec:training_stage}
We next describe the three training stages designed to progressively distill scene-graph reasoning into our model. A detailed breakdown of these datasets, including data formatting and distribution specifics, can be found in Appendix~\ref{sec:training_dataset}.

\paragraph{Stage 0: Image-segmentation-text alignment}
We adopt the training strategies from Osprey \cite{yuan2024osprey} to align the vision encoder, mask encoder, and language model. The vision encoders remain frozen throughout training, while the image-level projection layers, mask encoders, and the language model are trained in three progressive phases. First, we use LLaVA-Pretrain-558K image-text pairs \cite{zhang2023llava} to train only the  image-level projector. Second, we unfreeze the mask encoder to learn mask embeddings alongside the language model, focusing on object-level datasets (COCO \cite{coco}, RefCOCO, and RefCOCO+ \cite{yu_modeling_2016}) and part-level datasets (Pascal Part \cite{Chen2014DetectWY} and PartImageNet \cite{he2022partimagenetlargehighqualitydataset}). Finally, we finetune the language model on instruction tuning datasets for visual grounding and region understanding tasks, including Osprey-724K \cite{yuan2024osprey}, Visual Genome \cite{krishna2017visual}, and VCR \cite{zellers2019recognition}. 
This pre-training stage consists of 1.28M examples in total.

\paragraph{Stage 1: Instruction tuning with scene graphs}

From here, we unfreeze the visual encoder and train the model on three core data categories: (1) visual instructions, (2) grounding, and (3) scene graphs. Visual instructions consist of diverse visual instruction tasks including visual question answering, conversation, dense captioning, and OCR. For grounding, we incorporate referring expression comprehension~\cite{yu_modeling_2016} and region captions from Visual Genome~\cite{krishna2017visual}. The scene graph data includes PSG~\cite{yang2022panoptic}, Visual Genome \cite{krishna2017visual}, and our own \dataset-Stage1. The total number of examples in the Stage 1 training dataset is 1.73M. We refer to the resulting model trained with this mixture as \model-3B (Stage 1). Details of this data mixture are shown in Table~\ref{tab:stage1_dataset}.



\paragraph{Stage 2: Distillation with GPT4 edited scene graphs}
In this stage, we follow the methodology in Sec.~\ref{sec:dataset_stage2} and use \model-3B (Stage~1) to generate complete scene graphs, then apply our \distill pipeline with GPT-4 to build  \datasetsecond. We replace \datasetfirst with \datasetsecond, remove the OCR-focused portion of the visual-instruction data, and continue training \model-3B (Stage 1) on these updated inputs with 1.23M samples. This results in our final \model-3B. Details of this data mixture are shown in Table~\ref{tab:stage2_dataset}.

\section{Experiments}\label{sec:exp}

\begin{table*}[htb!]
\normalsize
\centering
\tablestyle{6pt}{1.0} 
\resizebox{\linewidth}{!}{%
\begin{tabular}{l l | c c c | c c c c}
\toprule
\multirow{2}{*}{\textbf{Model}} & \multirow{2}{*}{\textbf{LLM}} & \multicolumn{1}{c}{\textbf{GQA} \cite{hudson2019gqa}} & \multicolumn{1}{c}{\textbf{MMBench} \cite{liu2023mmbench}} & \textbf{SEED} \cite{li2023seedbenchbenchmarkingmultimodalllms} & \textbf{VSR} \cite{liu2023visual}  & \multicolumn{1}{c}{\textbf{CRPE} \cite{wang2024allseeing}} & \textbf{SugarCrepe} \cite{hsieh2024sugarcrepe} & \multicolumn{1}{c}{\textbf{What's Up?} \cite{kamath2023s}}  \\
 &  &  & \textbf{Dev-EN} & \textbf{Image} & \textbf{ZS-test} & \textbf{Relation} & \textbf{Relation} & \textbf{Controlled}  \\
\midrule
VILA1.5-3B \cite{lin2024vila} & ShearedLLaMA-2.7B & 61.5 & 63.4 & 67.9 & 61.0  & 67.8 & 86.3 & 50.6   \\
MiniCPM-V2.0-3B \cite{yao2024minicpm}  & Mini-CPM-2.4B &  & 69.7 & 67.1 & 68.2 &  68.1 & 86.6 & 54.8  \\
InternVL2-2B \cite{cai2024internlm2} & InternLM-2B & 61.0 & 73.2 & 70.9 & 69.0 & 65.8 & 85.5 & 74.4 \\
Phi-3-Vision \cite{abdin2024phi} & Phi-3-mini-4B & - & 74.2 & 71.0 & 67.8 & 71.6 & 88.7 & 78.7   \\
BLIP-3-single-image \cite{xue2024xgen} & Phi-3-mini-4B & - & 76.0 & 71.8 & 72.5 & 72.4 & 89.0 & 78.2 \\
BLIP-3-interleave \cite{xue2024xgen} & Phi-3-mini-4B & - & 76.8 & 72.2 & 72.6 & \textbf{72.5} & 88.3 & 76.3 \\
MM1-3B \cite{mckinzie2024mm1} & MM1-3B & - & 67.8 & 68.8 & - & - & - & -\\
MM1.5-3B \cite{zhang2024mm1} & MM1-3B & - & - & \textbf{72.4} & - & - & - & -\\
\midrule
\model-3B (Stage 1)  & Qwen2.5-3B & 60.8 & 77.0 & 70.6 & 73.7 & 65.9 & 89.1 & 81.3 \\
\model-3B & Qwen2.5-3B & 61.6 & \textbf{77.6} & 70.7 & \textbf{76.4} & 68.2 & \textbf{90.1} & \textbf{86.2} \\
\bottomrule
\end{tabular}
}
\caption {
    \small
    \textbf{Relationship understanding performance}: \model-3B (ours) vs. multimodal models with similar parameter size ($\leq$ 4B). Bold values show the best results.
}
\label{tab:relation_qa}
\vspace{0.5mm}
\end {table*}

In our experiments, we  compare \model-3B with the state-of-the-art (SoTA) open-source MLMs on visual reasoning tasks focused on relationship understanding. To see the performance gain from \distill, we separately evaluate \model-3B (Stage 1) and compare it against our final model, \model-3B. Additionally, we assess whether incorporating scene graph training enhances the model's capabilities in grounding and region understanding. Lastly, we directly evaluate \model-3B on scene graph generation to show its ability to produce accurate scene graphs. Prompts used for each task and dataset are provided in Appendix~\ref{sec:eval_details}.

\subsection{Relationship understanding benchmarks}
We first evaluate our model on a suite of visual question answering benchmarks specifically designed to probe relationship understanding capabilities. These benchmarks include: GQA~\cite{hudson2019gqa}, originally developed to assess relational understanding using image scene graphs; Visual Spatial Reasoning (VSR)\cite{liu2023visual}; MMBench\cite{liu2023mmbench}; SeedBench~\cite{li2023seedbenchbenchmarkingmultimodalllms}; CRPE~\cite{ma2023crepe}; SugarCrepe~\cite{hsieh2024sugarcrepe}; and What's Up~\cite{kamath-etal-2023-whats}. MMBench and SeedBench are selected for their comprehensive coverage of spatial relations and object interactions. The CPRE dataset focuses on relation comprehension that also includes abnormal relations generated with synthetic images. We report the overall average, denoted as "Relation." For SugarCrepe, we use the ``replace-relation" split and follow their binary multiple-choice evaluation framework suitable for MLMs. For What's Up, we evaluate on 820 images containing unambiguous object relationships captured in controlled environments (denoted as "Controlled").


Table~\ref{tab:relation_qa} shows the performance comparison of \model-3B with models of similar size ($\leq$4B parameters)~\cite{lin2024vila, yao2024minicpm, cai2024internlm2, abdin2024phi, xue2024xgen, zhang2024mm1}. We observe that \model-3B achieves the strongest performance on MMBench, VSR, SugarCrepe, and Whats Up, while maintaining competitive performance on others. Notably, despite being trained on fewer than 3M instances, \model-3B surpasses Phi-3-Vision and BLIP-3 on the What's Up dataset by a significant margin (86.2\% vs. 78.7\% and 78.2\%, respectively). The same is observed in VSR where \model-3B greatly exceeds BLIP-3 (76.4\% vs. 72.6\%), showcasing our model's strengths in understanding spatial relations in general. This is particularly impressive given that Phi-3-Vision and BLIP-3 have undergone extensive pre-training on up to 300 million instances. 

Comparing our Stage 1 and Stage 2 models, we find that while performance remains stable on MMBench and SeedBench, the Stage 2 model shows consistent improvements across relationship understanding benchmarks. We observe substantial gains in CRPE (68.2\% vs 65.9\%), SugarCrepe (90.1\% vs 89.1\%), and Whats Up (86.2\% vs 81.3\%). This demonstrates the effectiveness of our GPT-4 distillation approach for enhancing relational reasoning.



\begin{table*}[ht!]
\normalsize
\centering
\tablestyle{11pt}{0.98} 
\resizebox{\textwidth}{!}{

\begin{tabular}{l | c c c  c c c  c c | c c}

\toprule
\multirow{2}{*}{\textbf{Model}} & \multicolumn{3}{c}{\textbf{RefCOCO} \cite{yu2016modeling}} & \multicolumn{3}{c}{\textbf{RefCOCO+} \cite{yu2016modeling}} & \multicolumn{2}{c|}{\textbf{RefCOCOg} \cite{mao2016generation}} & \multirow{2}{*}{\textbf{Avg}} & \multirow{2}{*}{\textbf{Avg\textsubscript{test}}} 
\\
 & Val & Test-A & Test-B & Val & Test-A & Test-B & Val & Test &  & \\
\midrule
Shikra-7B \cite{chen2023shikra} & 87.0 & 90.6 & 80.2 & 81.6 & 87.4 & 72.1 & 82.3 & 82.2 & 82.9 & 82.5\\
MiniGPT-v2-7B  \cite{chen2023minigpt} & 88.1 & 91.3 & 84.3 & 79.6 & 85.5 & 73.3 & 84.2 & 84.3 & 83.8 & 83.7 \\
QwenVL-7B \cite{bai2023qwen} & 88.6 & 92.3 & 84.5 & 82.8 & 88.6 & 76.8 & 86.0 & 86.3 & 85.7 & 85.7 \\
Ferret-7B \cite{you2023ferret} & 87.5 & 91.3 & 82.4 & 80.8 & 87.4 & 73.1 & 83.9 & 84.8 & 83.9 & 83.8 \\
Groma-7B \cite{ma2025groma} & 89.5 & 92.1 & 86.3 & 83.9 & 88.9 & 78.1 & 86.4 & 87.0 & 86.5 & 86.5 \\
VisionLLMv2-Chat \cite{wu2024visionllm} & 90.0 & 93.1 & 87.1 & 81.1 & 87.3 & 74.5 & 85.0 & 86.4 & 85.6 & 85.7 \\
Ferret-13B \cite{you2023ferret} & 89.5 & 92.4 & 84.4 & 82.8 & 88.1 & 75.2 & 85.8 & 86.3 & 85.6 & 85.3\\
ASM-V2-13B \cite{wang2024allseeing} & 90.6 & 94.2 & 86.2 & 84.8 & 90.8 & 76.9 & 87.5 & 88.3 & 87.4 & 87.3 \\
\midrule 
InternVL2-2B \cite{cai2024internlm2} & 82.3 & 88.2 & 75.9 & 73.5 & 82.8 & 63.3 & 77.6 & 78.3 & 77.8 & 77.7 \\
Phi-3-Vision-4B \cite{abdin2024phi}  & - & 46.3 & 36.1 & - & 42.0 & 28.8 & - & 37.6 & - & 38.1 \\ 
MM1.5-3B \cite{zhang2024mm1} & - & 92.0 & 86.1 & - & 87.7 & 75.9 & - & 86.4 & - & 85.6 \\
Qwen2.5-VL-3B~\cite{bai2025qwen25vltechnicalreport} & 89.1 & 91.7 & 84.0 & 82.4 & 88.0 & 74.1 & 85.2 & 85.7 & 85.0 & 84.7 \\


\model-3B (Stage 1) & 90.0 & 93.6 & 87.2 & 84.6 & 90.5 & 77.7 & 87.2 & 87.1 & 87.2 & 87.2 \\
\model-3B & 91.6 & 94.3 & 88.6 & 86.8 & 91.8 & 80.5 & 88.5 & 88.8 & \textbf{88.9} & \textbf{88.8} \\

\bottomrule
\end{tabular}
}
\caption {
\small
Comparison with SoTA models up to 13B parameters on \textbf{Referring Expression Comprehension}. The results are reported based on Recall@1 with IoU $>$ 0.5. 
} 
\label{tab:results_rec}
\vspace{0.5mm}
\end {table*}

\begin{table}[h!]
\centering
\tablestyle{2pt}{0.98} 
\resizebox{\linewidth}{!}{%
\begin{tabular}{l|ccc|cc cc}

\toprule
\multirow{3}{*}{\textbf{Model}} & \multicolumn{3}{c|}{\textbf{Open-Vocab. Segmentation}} & \multicolumn{4}{c}{\textbf{Region Classification}} \\
\cmidrule(lr){2-4} \cmidrule(lr){5-8}
& \multicolumn{3}{c|}{\textbf{ADE} \cite{zhou2019semantic}} & \multicolumn{2}{c}{\textbf{LVIS} \cite{gupta2019lvis}} & \multicolumn{2}{c}{\textbf{PACO} \cite{ramanathan2023paco}} 
\\
 & PQ & mAP & mIoU & SS & S-IOU & SS & S-IOU \\
\midrule
LLaVA-1.5-7B \cite{zhang2023llava} & - & - & - & 48.9 & 19.8 & 42.2 & 14.6 \\
Kosmos-2 \cite{peng2023kosmos} &6.5 & 4.3 & 5.4 & 38.9 & 8.7 & 32.1 & 4.8 \\
Shikra-7B \cite{chen2023shikra} & 27.5 & 20.3 & 18.2 &  49.6 & 19.8 & 43.6 & 11.4 \\
GPT4RoI-7B \cite{zhang2023gpt4roi} & 36.3 & 26.1 & 25.8 & 51.3 & 12.0 & 48.0 & 12.1 \\
Ferret-7B \cite{you2023ferret} & 39.5 & 29.9 & 31.8 & 63.8 & 36.6 & 58.7 & 26.0 \\
Osprey-7B \cite{yuan2024osprey} & 41.9 & 41.2 & 29.6 & 65.2 & 38.2 & 73.1 & 52.7 \\
VisionLLMv2-Chat \cite{wu2024visionllm} & - & - & - & 67.3 & 42.7 & 63.8 & 36.3\\
\midrule

\modelname-3B [Stage 1] & 41.1 & 40.3 & 30.2 & 71.2 & 47.6 & 71.5 & 50.3 \\
\modelname-3B & \textbf{44.2} & \textbf{45.2} & \textbf{33.9} & \textbf{72.6} & \textbf{49.8} & \textbf{74.1} & \textbf{53.1} \\
\bottomrule
\end{tabular}
}
\caption {
Results on \textbf{Region Recognition}. Following\cite{yuan2024osprey}, we report panoptic segmentation (PQ), instance segmentation (mAP), and semantic segmentation (mIoU) on the ADE20K validation set. For region classification, we measure referring object classification on object-level LVIS and part-level PACO, reporting Semantic Similarity (SS) and Semantic Intersection over Union (S-IOU). 
} 
\label{tab:results_region_recognition}
\vspace{0.5mm}
\end {table}
\subsection{Referring expression comprehension}
We next assess the grounded reasoning capabilities of our model on referring expression comprehension tasks using the RefCOCO, RefCOCO+, and RefCOCOg datasets \cite{yu2016modeling, mao2016generation}. We prompt our models to provide a bounding box for each description, and report Recall@1 (IoU $>$ 0.5). As shown in Table~\ref{tab:results_rec}, \modelname-3B achieves the highest overall average accuracy of 88.8\% on the test split, surpassing larger models such as ASM-V2-13B (87.3\%). We significantly outperform MM1.5-3B (85.6\%) that has been trained with more than 1M instances for grounding. Our model also exceeds Qwen2.5-VL-3B~\cite{bai2025qwen25vltechnicalreport} which previously demonstrated powerful visual reasoning capabilities among similar sized models. Finally, the improvement from Stage 1 to Stage 2 (87.2\% $\rightarrow$ 88.8\%) further validates the effectiveness of our self-distillation approach for enhancing grounded reasoning. These results demonstrate that incorporating scene graph training significantly improves the grounding capabilities, allowing our model to surpass models with significantly more parameters and training data.

\subsection{Region recognition}
We evaluate our model on open-vocabulary region recognition tasks, namely semantic segmentation on the ADE20k dataset \cite{zhou2019semantic} and region classification on the LVIS and PACO datasets \cite{ramanathan2023paco, gupta2019lvis}. For each task, the model generates a description or category label for the region specified by a segmentation mask and bounding box; we then convert text outputs to class labels using SentenceBERT~\cite{reimers2019sentence} similarity, following \cite{yuan2024osprey}. 

Table~\ref{tab:results_region_recognition} shows the results. Overall, our \model-3B outperforms up to 7B-scale models across all metrics. On ADE20K open-vocabulary segmentation task, we exceed the best-performing Osprey-7B by +2.3\% PQ, +4.0\% mAP, and +2.3\% mIoU. Compared to VisionLLMv2-7B, our model improves LVIS by +5.3\% SS and +7.1\% S-IOU and PACO by +10.3\% SS and +16.8\% S-IOU. Finally, progressing from Stage1 to Stage2 again helps across the board with additional gain of +2.6\% SS and +2.8\% S-IOU on PACO, which allows \model-3B to surpass Osprey-7B on this dataset.



\subsection{Scene graph generation}

Lastly, we evaluate \model{} on the task of scene graph detection using the Panoptic Scene Graph (PSG) dataset \cite{yang2022panoptic}. Here, the model must generate bounding boxes for object regions of interest and provide subject-predicate-object triplets for these detected regions. 
We prompt \model to produce a complete scene graph, including bounding boxes, parse the output to extract object regions and predicates, and assign the text outputs to the closest object and predicate labels in the dataset via semantic similarity. We report Recall@20 (R@20) and mean Recall@20 (mR@20), which measure whether the ground-truth triplets appear in the top $K=20$ predictions and match in bounding box (IoU $>$ 0.5) and class labels.


We compare against MLMs that generate scene graphs as open-ended text \cite{wang2024allseeing,zhao2023textpsg},  as well as existing PSG detection-based models  \cite{xu2017scene,zellers2018neural,tang2019learning,lin2020gps,yang2022panoptic}, which have been single-task fined-tuned on the PSG dataset to classify the relation triplets from a pre-defined closed set of class labels. Table~\ref{tab:results_psg} shows that our \model-3B (Stage~1) model already outperforms previous open-ended generation models like ASM-V2-13B (18.7/12.0 vs. 14.2/10.3 on R@20/mR@20) After applying scene graph self-distillation in Stage~2, our \model-3B model further improves to \textbf{21.0/13.2 on R@20/mR@20}. This demonstrates that scene graph self-distillation helps our model generate more accurate and diverse relation predictions. When comparing to closed-set detection models, our \model-3B is on par with early closed-set models like MOTIFS and VC Tree, both reporting an R@20 of around 20.0. Notably, these closed-set models are single-task fine-tuned on the PSG dataset and operate within a constrained set of relation classes, whereas our model operates in an open-ended setting without pre-defined set of relation classes. Overall, these results suggest the effectiveness of our approach in generating accurate scene graphs in an open-ended manner. 

\begin{table}[ht!]
\normalsize
\centering
\tablestyle{8pt}{0.98} 
\resizebox{\linewidth}{!}{%
\begin{tabular}{l |  c c c }

\toprule
\textbf{Method} & \# Relations & R@20 & mR@20 \\ 
\midrule
\textit{Open-Ended Generation Models} \\
\midrule
TextPSG \cite{zhao2023textpsg}   & 50.0 & 4.8 & - \\
ASM-V2-13B \cite{wang2024allseeing} & 9.2 & 14.2 & 10.3 \\

\model-3B (Stage1) & 5.6 & 18.7 & 12.0 \\
\model-3B & 6.1 & \underline{21.0} & \underline{13.2} \\



\midrule
\textit{Closed-Set Detection Models} \\
\midrule
IMP \cite{xu2017scene} & 20.0 &  16.5 & 6.5 \\ 
MOTIFS \cite{zellers2018neural} & 20.0 & 20.0 & 9.1 \\ 
VC Tree \cite{tang2019learning} & 20.0 & 20.6 & 9.7 \\
GPSNet \cite{lin2020gps} & 20.0 & 17.8 & 7.0 \\
PSGFormer \cite{yang2022panoptic} & 20.0 & 18.6 & 16.7 \\
HiLO \cite{zhou2023hilo} & 20.0 & 40.6 & 29.7 \\
DSGG \cite{hayder2024dsgg} & 20.0 & 36.2 & 34.0 \\
\bottomrule
\end{tabular}
}
\caption {
\small
\textbf{Panoptic Scene Graph (PSG)} generation results. We report the recall (R@20) and mean recall (mR@20) of the predicted triplet relations. Underlined values denote the best results among open-ended generation models. Note that all closed-set models are single-task fine-tuned on the PSG dataset.
} 
\label{tab:results_psg}
\vspace{0.5mm}
\end {table}

\subsection{Ablation studies} 
In our ablation studies, we focus on showing the effectiveness of our proposed instruction tuning with dense scene graphs, and self-distillation with refinement from GPT4.

\paragraph{Role of scene graphs in instruction tuning}
To understand the benefits of incorporating scene graph data in visual understanding, we trained model variants using different combinations of the three datasets introduced in Stage~1 training (Sec~\ref{sec:training_stage}). We excluded VSR from the Visual Instruction (Vis-Ins) data to avoid train-test overlap and ensure fair zero-shot evaluation of relationship reasoning capabilities. Table~\ref{tab:sg_ablations} shows the results comparing models trained with and without scene graph data. 
In region classification, we observe that adding scene graph (SG) data provides consistent gains across different training data configurations. Interestingly, the highest performance on the PACO dataset is achieved when training with only visual instruction and SG data (72.5 in SS). We suspect this is because the RefCOCO dataset is biased towards COCO objects rather than fine-grained, part-level objects. In contrast, our scene graph data encompass diverse relationships including part-level objects, providing enhanced reasoning capabilities for this task. Similarly, in relationship understanding tasks, we observe the same benefits of adding SG data with gains of 1.5\% in VSR and 0.7\% in CRPE.

\begin{table}[h!]
\normalsize
\centering
\vspace{-0.5em}
\tablestyle{4pt}{0.98} 
\resizebox{\linewidth}{!}{%
\begin{tabular}{ c c c | c  c | c c}
   \toprule
   \multicolumn{3}{c|}{\textbf{Training Dataset}}  & \multicolumn{2}{c|}{\uline{\textbf{Region Classification}}} & \multicolumn{2}{c}{\uline{\textbf{Relationship Understanding}}}  \\
   & & &  LVIS & PACO & VSR & CRPE \\
   Grounding & Vis-Ins & SG  & SS  & SS & ZS-test &  Relation \\
   \midrule
  - & - & - & 50.3 & 48.0 & 48.6 & 51.6 \\
   \checkmark & - & -   & 49.3 & 47.1 & 48.1 & 54.0 \\
   \checkmark & - & \checkmark & 55.3 & 53.5 & 48.9 & 60.4 \\
   - & \checkmark & -  & 65.4 & 68.9 & 69.1 & 63.5\\
   - & \checkmark & \checkmark  & 70.5 & \textbf{72.5} & 69.3 & 64.1\\
   \checkmark & \checkmark & -  & 68.6 & 70.5 & 68.2 & 64.1 \\
   \checkmark & \checkmark & \checkmark  & \textbf{70.9} & 71.1 & \textbf{69.7} & \textbf{64.8} \\
   \bottomrule
   \end{tabular}
}
\caption{
 Ablation studies of instruction tuning with scene graph data in Stage 1. Different models trained with combinations of visual instruction (Vis-Ins),  grounding (Grounding), and scene graph (SG) data, which are categorized in Stage 1 training (Section~\ref{sec:training_stage}), are evaluated on region classification (SS: semantic similarity) and relationship understanding tasks. In this study, we focus on measuring true zero-shot performance and remove  VSR~\cite{liu2023visual} from the Vis-Ins training set.  Note, the last row corresponds to the Robin-3B (Stage 1) model excluding VSR.
  }
\label{tab:sg_ablations}
\end {table}

\begin{figure}[ht!]
    \centering
    \includegraphics[width=0.9\linewidth]{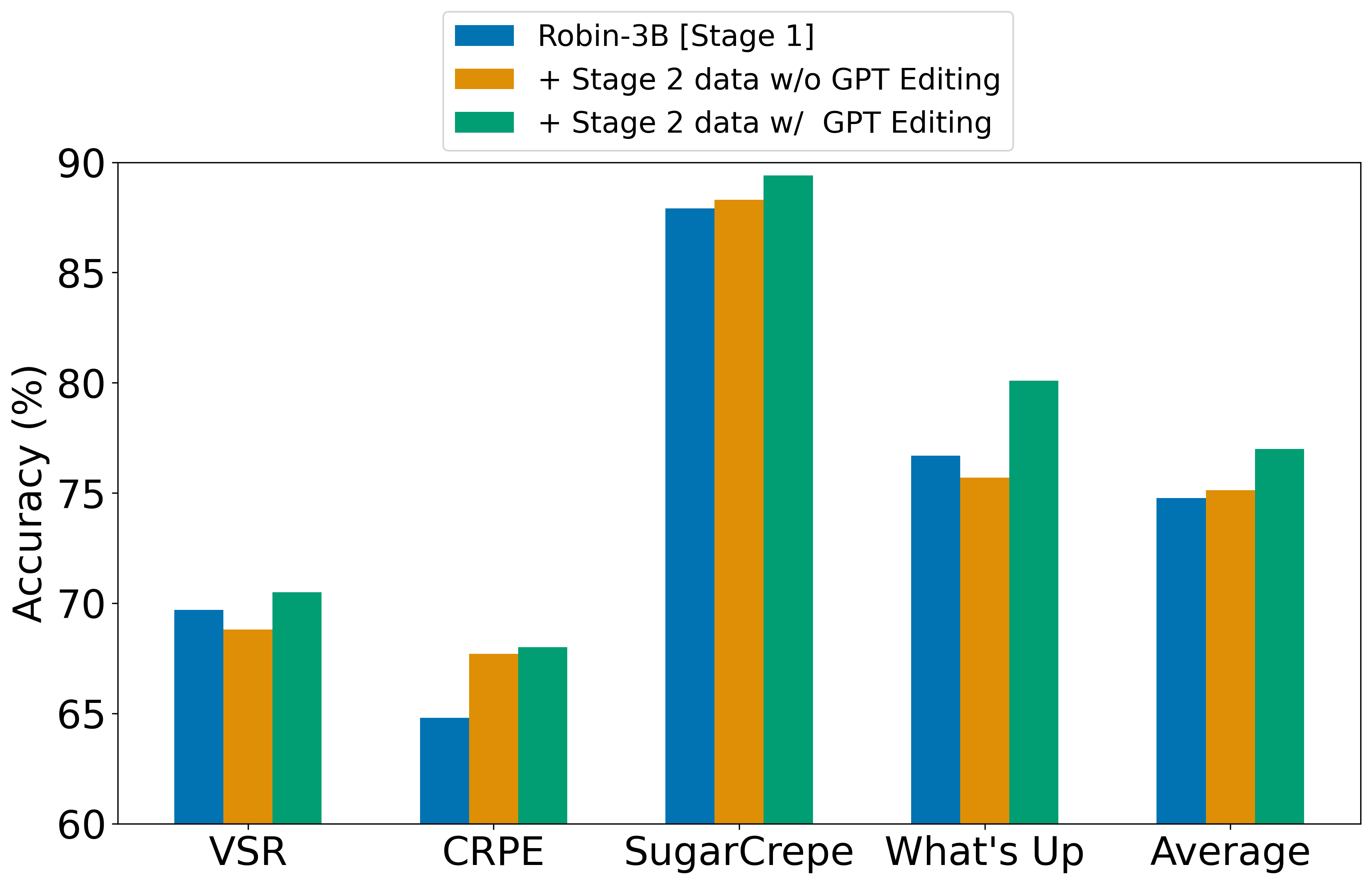}
    \vspace{-1mm}
    \caption{Effectiveness of \distill distillation.}
    \label{fig:stage2_ablation}
\end{figure}

\paragraph{Self-distillation and GPT4 editing}
Based on our \distill framework, we investigate the empirical gains of using GPT-4 edited scene graphs compared to training on the model's own generated scene graphs without refinement. Figure~\ref{fig:stage2_ablation} compares the performance of \model-3B trained with and without the GPT-4 edited scene graphs on relationship understanding benchmarks (with VSR again excluded from training). We see a marginal average improvement of 0.5\% over the Stage~1 model, when training the model additionally with its own scene graph generations (w/o GPT Editing). This gain is primarily due to improvements in the CRPE (+2.9\%) and SugarCrepe (+0.4\%) datasets, whereas performance decreases on the VSR (-0.9\%) and What's Up (-1.0\%) datasets. Meanwhile, we observe consistent gains for all benchmarks using edited scene graph (w/ edit) with average improvement of 2.2\%.



\begin{figure}[t]
    \centering
    \includegraphics[width=0.8\linewidth]{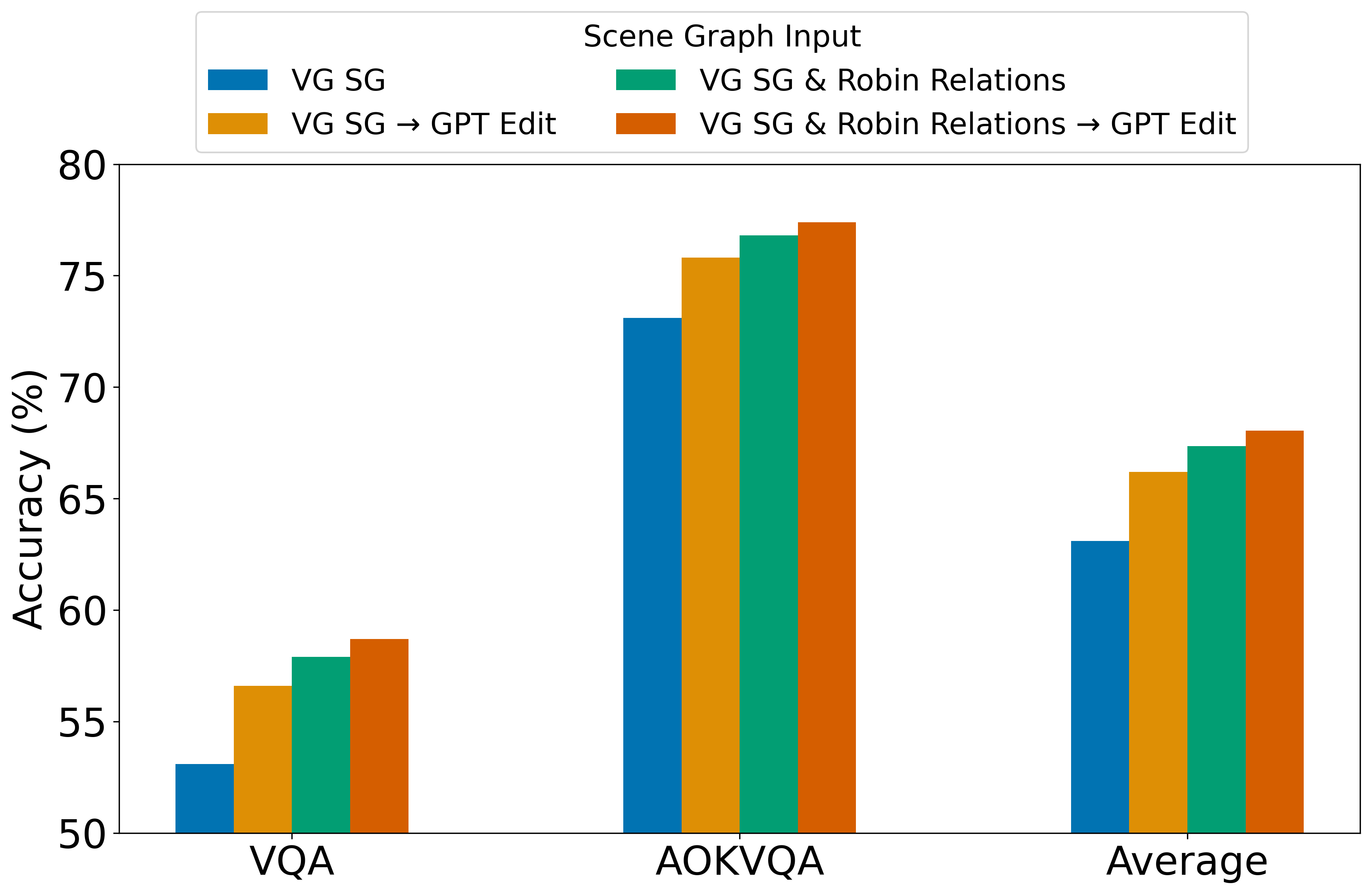}
    \vspace{-1mm}
    \caption{
    VQA performance of blind LLMs using different scene graphs as context: (1) original Visual Genome (VG) scene graphs, (2) GPT-edited VG scene graphs, (3) VG scene graphs appended with \modelname-3B relations (Stage 1), and (4) the GPT-edited scene graphs from Stage 2.
    }
    \label{fig:scene_graph_eval}
\end{figure}

\paragraph{Human-annotated vs. machine edited scene graphs}
Given the consistent gains from GPT-4 edited scene graphs, a natural question arises: can we simply refine human-annotated scene graphs with GPT-4 instead of relying on model-generated graphs?
We hypothesize that a high-quality scene graph must encode sufficient visual details for a language model to accurately answer questions. To test this, we run an ablation study on VQAv2~\cite{vqav2} and AOKVQA~\cite{AOKVQA} in a “blind LLM” setting, where the model has no direct image input but only a scene graph. As Figure~\ref{fig:scene_graph_eval} shows, editing scene graphs with GPT-4 (\textrightarrow{} GPT Edit) consistently improves VQA performance compared to unedited versions. Notably, the model-generated scene graphs (VG SG \& \modelname Relations) outperform GPT-4 edits on human annotations alone (VG SG~\textrightarrow GPT Edit), indicating that our model captures additional relationships missing from Visual Genome. Finally, applying GPT-4 edits on these model-generated graphs (VG SG \& \modelname Relations\textrightarrow~GPT Edit) leads to the highest accuracy, validating the proposed framework.

\section{Qualitative Results}

\begin{figure*}[ht!]
    \centering
    \begin{subfigure}[t]{0.84\textwidth}
        \centering
        \includegraphics[width=\textwidth]{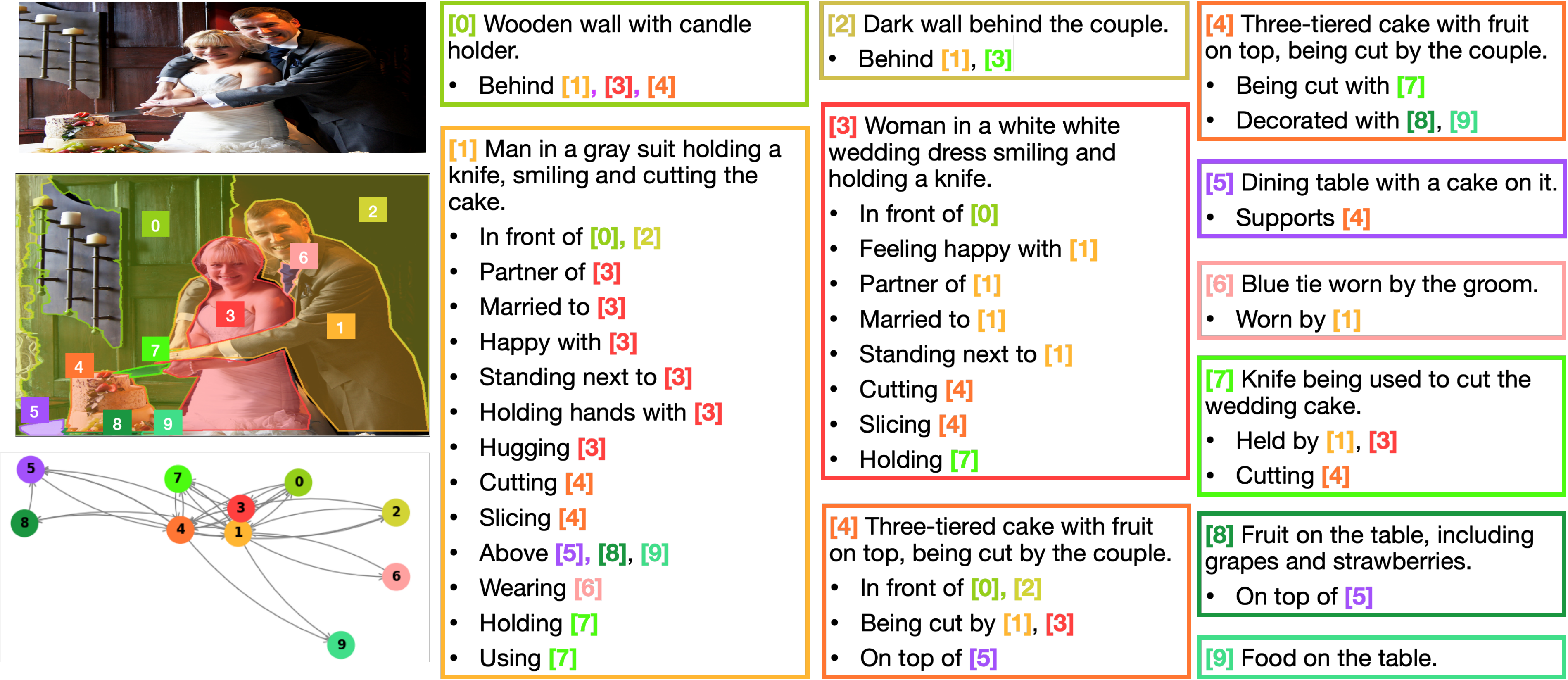}
        \caption{}
        \label{fig:qual_ex1}
    \end{subfigure}
    \vskip\baselineskip  
    \begin{subfigure}[t]{0.84\textwidth}
        \centering
        \includegraphics[width=\textwidth]{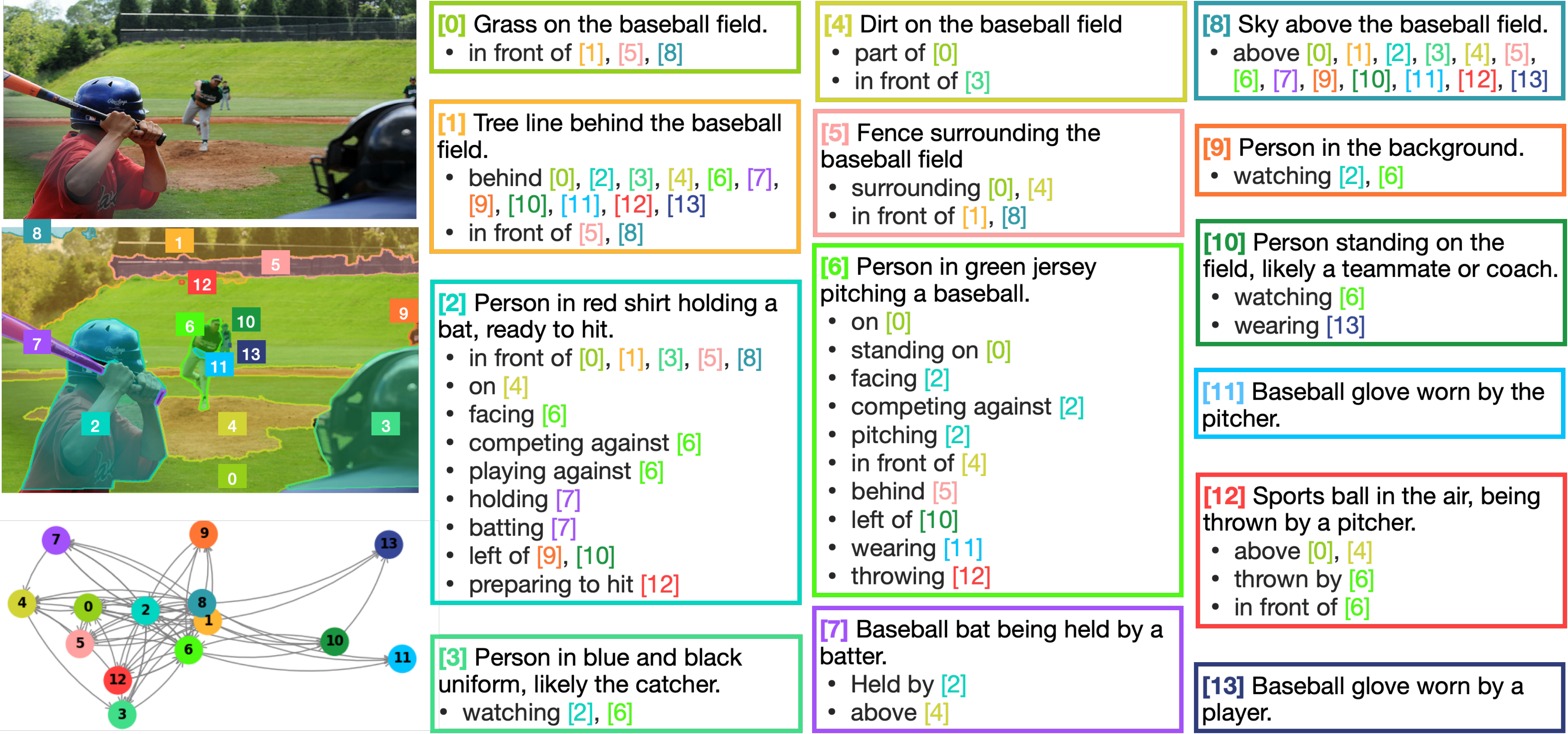}
        \caption{}
        \label{fig:qual_ex2}
    \end{subfigure}
    \vskip\baselineskip
    \caption{
    Dense scene graph generated by \model\ on the Panoptic Scene Graph dataset~\cite{yang2022panoptic}}
    \label{fig:robin_sg_psg}
\end{figure*}

\begin{figure*}[ht!]
    \centering
    \includegraphics[width=0.95\textwidth]{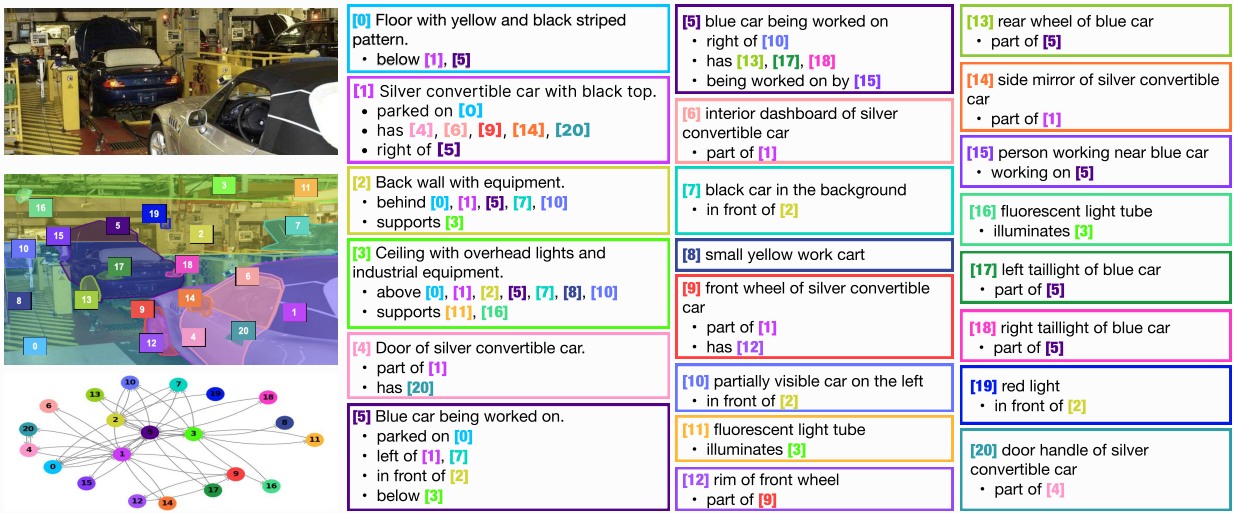}
    \caption{
    Dense scene graph generated by \model\ on the ADE20K dataset~\cite{zhou2019semantic}
    }
    \label{fig:robin_sg_ade}
\end{figure*}

We present qualitative results demonstrating \modelname's ability to generate dense scene graphs from segmentation masks. Our examples use ground-truth masks from the Panoptic Scene Graph (PSG) dataset~\cite{yang2022panoptic} and ADE20K dataset~\cite{zhou2019semantic}, though our framework can leverage any segmentation generator (e.g., SAM~\cite{kirillov2023segment}) to produce dense scene graphs for arbitrary segmented scenes.

Figure~\ref{fig:robin_sg_psg} showcases how \model captures a diverse set of relationships among the objects in the scene.
In Figure~\ref{fig:qual_ex1}, the relationships associated with the man in a gray suit ([1]) include spatial (e.g., \textit{in front of} [0], [2]; \textit{standing next to} [3]), social (e.g., \textit{partner of} [3], \textit{married to} [3]), emotional (e.g., \textit{happy with} [3]), and interactional (e.g., \textit{hugging} [3], \textit{cutting} [4], \textit{holding} [7]). 
This richness surpasses previous classification-based scene graph models, capturing the comprehensive relationships involving the man in the scene. In addition, the generated descriptions go beyond conventional object labels, providing detailed object descriptions helpful for understanding the scene, such as the cake [4] being a ``three-tiered cake with fruit on top, being cut by the couple'', or the ``wooden wall with candle holder'' [0]. The model also demonstrates precise spatial reasoning-correctly identifying that the dark wall [2] is "behind" the scene while excluding the cake [4] from this relationship.

Figure~\ref{fig:qual_ex2} presents another set of dense relationships. It accurately locates the batter [2] and the pitcher [6], including the relationship that they are competing against each other. People in the background ([9] and [10]) are accurately identified, and the model includes salient information that they are watching the two players ([2] and [6]). Notably, the model is able to identify small objects, such as the sports ball in the air [12], and correctly infers that it is thrown by the pitcher [6], and that the batter [2] is preparing to hit it [12], demonstrating strong localization ability enabled in pixel and text space.

Figure~\ref{fig:robin_sg_ade} demonstrates part-whole relationship understanding in complex scenes. \model accurately identifies vehicle components-dashboard [6], door [4], front wheel [9], and handle [20]-as parts of the "silver convertible car" [1].
It distinguishes spatial markers, such as the ``left taillight'' [17] and ``right taillight'' [18], as parts of the blue car [5]. Furthermore, it accurately associates the person [15] as working on the blue car [5] rather than other cars. This showcases \model can also achieve precise entity-relationship grounding in cluttered environments.

\section{Related work}

\textbf{Relationship understanding and scene graphs} 
Understanding relationships in visual scenes is a crucial challenge in computer vision, commonly addressed via scene graph generation (SGG)~\cite{xu2017scene}. Compared to visual question answering~\cite{hudson2019gqa,liu2023visual}, scene graphs require a more comprehensive and structured understanding of the objects and their relationships within an image. Datasets such as Visual Relationship Detection~\cite{lu2016visual} and Visual Genome~\cite{krishna2017visual} have stimulated research into SGG models, while Panoptic scene graph generation (PSG)~\cite{yang2022panoptic} has extended bounding boxes based SGG to panoptic segmentation.  Numerous end-to-end architectures rely on classification-based methods\cite{xu2017scene, zellers2018neural, yuan2022rliprelationallanguageimagepretraining, li2022sgtrendtoendscenegraph, zhao2023textpsg}, and efforts exist to address object-relationship biases~\cite{tang2020unbiased, zhou2023hilo} or to distill relationships from LLMs or MLMs~\cite{Kim2023LLM4SGGLL, chen2024scene, Li2024FromPT}. However, these distillation-based approaches focus on improving the SGG task itself rather than enhancing the underlying MLMs.

\textbf{Visual instruction tuning and MLMs}
The recent advancements in large language models~\cite{openai2024gpt4} have resulted in the emergence of numerous LLM-based multimodal models~\cite{zhu2023minigpt,li2023mimic,ye2023mplug,liu2023improved}. 
LLaVA is the first work that introduces visual instruction for LLMs, 
where the authors use language-only GPT-4 to generate a multi-modal instruction-following dataset 
~\cite{liu2024visual}. Follow-up works have built other types of multi-modal instruction tuning data, including conversational-style QA~\cite{zhu2023minigpt}, single-round QA based on academic datasets~\cite{dai2024instructblip}, detailed image descriptions~\cite{chen2023sharegpt4v}. 

Among the variants of MLMs, grounded MLMs~\cite{chen2023minigpt,bai2023qwen,wang2023cogvlm,zhao2023bubogpt,you2023ferret,rasheed2023glamm} are most relevant to our study, which are trained on region-based instruction tuning data to enable grounding capabilities of existing MLMs. A common representation of regions in such models is to refer the
object regions as their bounding box coordinates in the text input-output~\cite{zhang2023llava,chen2023shikra}. Alternatively, Osprey~\cite{yuan2024osprey} is designed to understand a single or a handful of object segmentations at different levels of granularity. In this work, we extend Osprey to scene graph generation for a more comprehensive understanding of the entire image. ASM-v2~\cite{wang2024allseeing} is a grounded MLM designed for relation understanding and scene graph generation.
Our approach relies on reasoning over relationships with diverse categories inferred by GPT4-V, and creates more complete and comprehensive scene graphs. 

\textbf{Data filtering of image-text datasets}
Radenovic et al. propose a rule-based system 
that filters out examples with low complexity 
\cite{radenovic2023filtering},  
while LAION-400M introduces CLIP-filtering to remove image-text pairs with low similarity \cite{schuhmann2021laion, clip}. Since then, various works have proposed improved filtering methods based on CLIP \cite{gadre2024datacomp,fang2023data}.
A more recent work reports that finetuning MLMs such as LLaVa can yield better image-text data filters \cite{wang2024finetuned}. In contrast, we have developed a novel filtering pipeline that combines rule-based and model-based filtering for image-scene-graph data. 


\section{Conclusion}

We present \dataset and \model to enhance the visual reasoning capabilities of MLMs.  Our model outperforms the SoTA approaches in a suit of relationship reasoning tasks, grounded reasoning, and open-ended panoptic scene graph generation. Using the \distill distillation framework, \model-3B model outperforms the state of the art 3B models in relationship understanding and REC tasks, and even 13B models trained with a similar amount of data. Future work should evaluate scene graph generations beyond existing annotations on COCO images to any image in the wild. Future work should also expand this work to predicting 3D shapes, extracting graphs from videos, and even facilitating image generation conditioned on scene graphs. 

\noindent\textbf{Acknowledgements.} This work is partially funded by Toyota Motor Inc. and partially by Amazon Science Hub.




\bibliographystyle{ieeenat_fullname}
\bibliography{references}

\newpage

\appendix
\onecolumn
\setcounter{page}{1}
\begin{center}
    \textbf{\LARGE Synthetic Visual Genome} \\
    \LARGE Supplementary Material
\end{center}


\section{Qualitative Analysis}
In this section, we provide qualitative results and analysis that validate our scene graph data generation pipeline and design rationale. 

\subsection{Stage 1 vs. Stage 2 Training}
\begin{figure*}[ht!]
    \centering
    \includegraphics[width=0.7\textwidth]{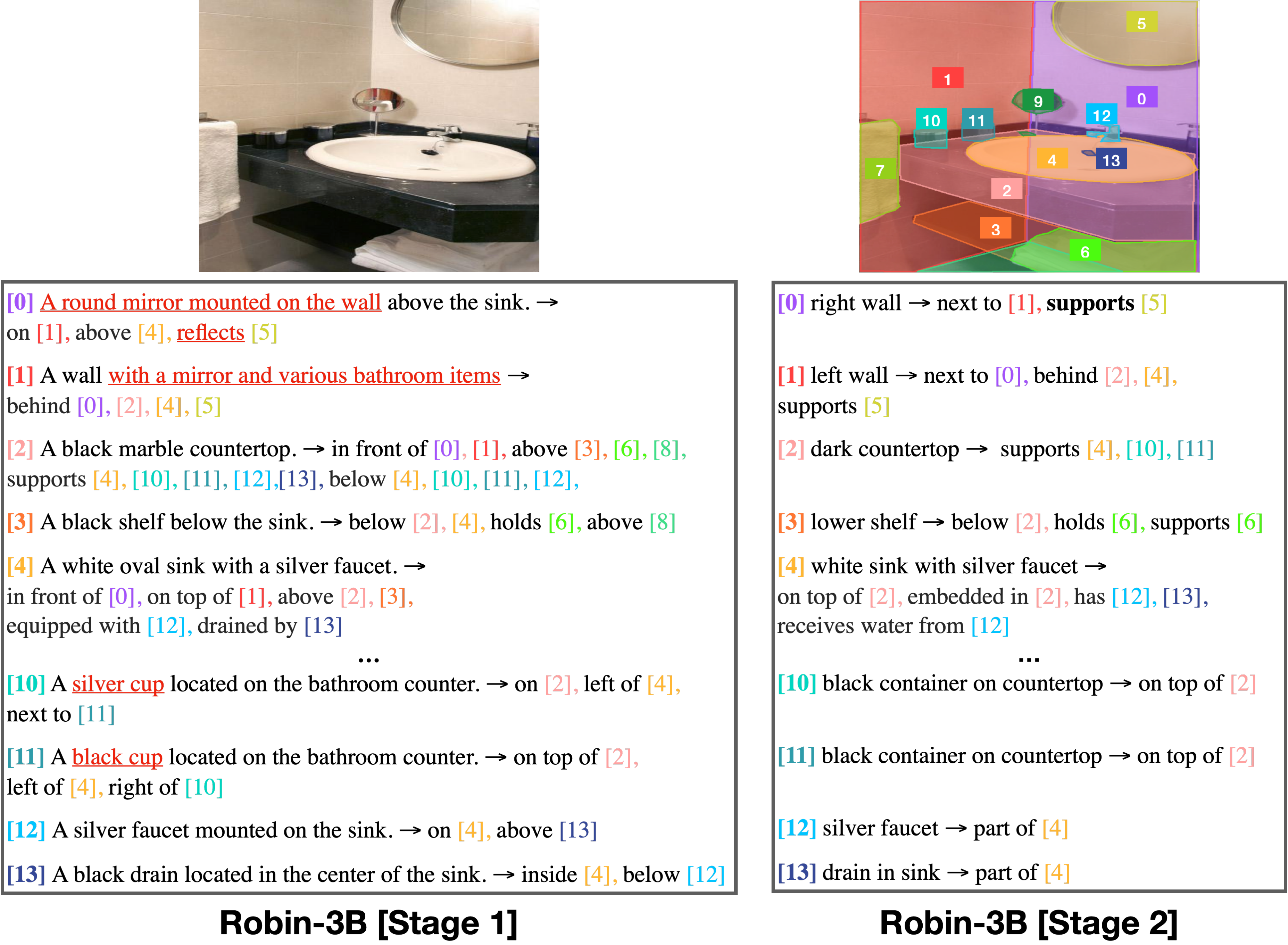}
    \caption{Comparison of models trained in Stage~1 and Stage~2. Errors are highlighted and underlined in red. The Stage~1 model lacks precise understanding of the segmentation mask, incorrectly mentioning a mirror for regions [0] and [1], and is more prone to hallucinations (e.g., mentioning a cup on regions [10] and [11]).}
    \label{fig:qual_stage1_stage2}
\end{figure*}

Figure~\ref{fig:qual_stage1_stage2} presents a qualitative comparison between the Stage~1 and Stage~2 trained versions of \model-3B. Notably, the Stage~2 model demonstrates a more accurate and precise understanding of the segmentation mask, correctly labeling [0] and [1] as the left and right walls, respectively. In contrast, the Stage~1 model inaccurately labels [0] and [1] as ``round mirror,'' and exhibits hallucinations by stating that there are cups on the bathroom counter in regions [10] and [11]. These observations indicate that Stage~2 training is crucial for resolving such issues and improving the visual perception capabilities of the model.

\subsection{GPT-4 Generated Scene Graph}
\label{sec:som_scene_graph}
\begin{figure*}[ht!]
    \centering
    \includegraphics[width=0.7\textwidth]{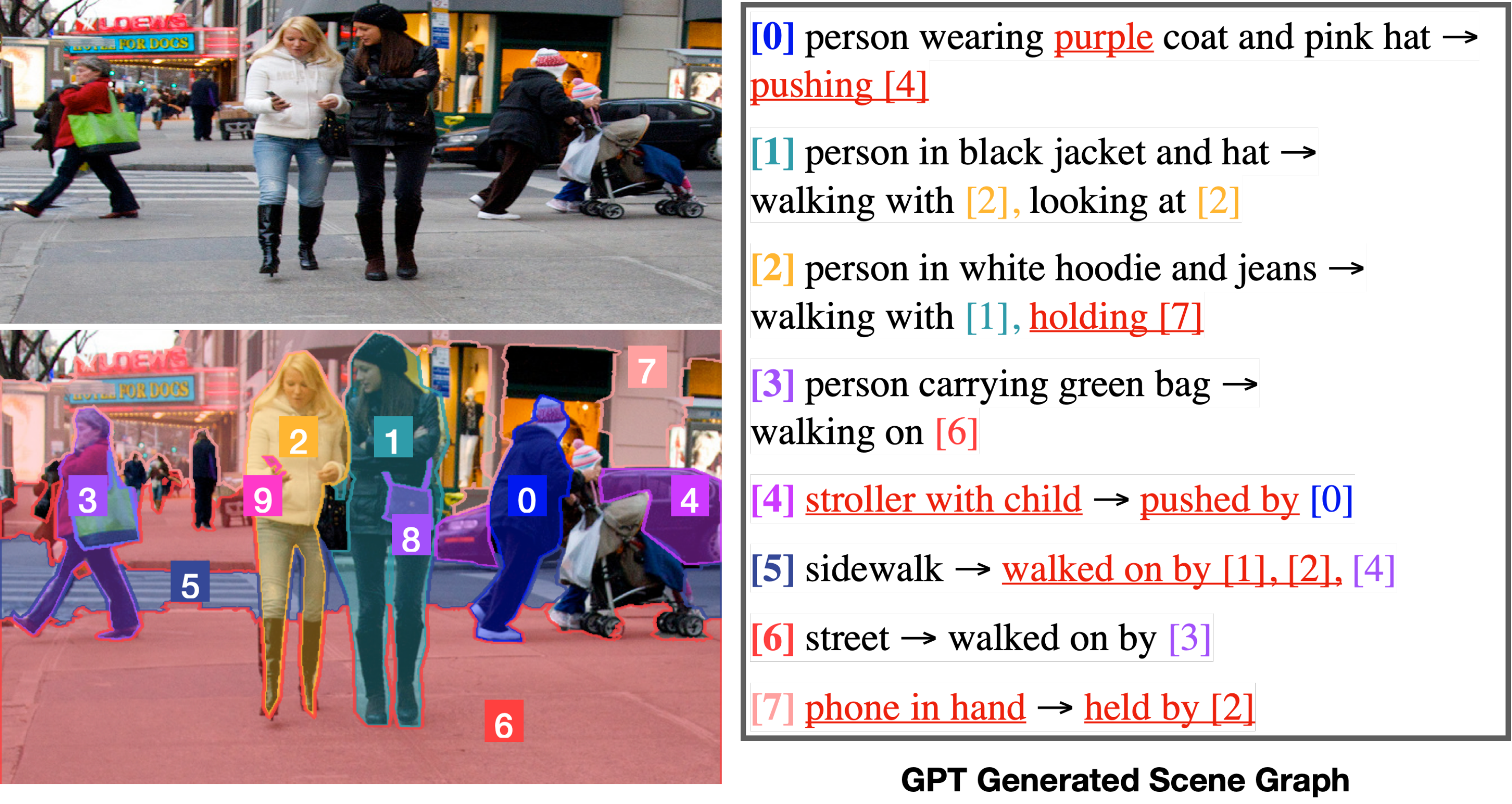}
    \caption{Example of scene graph quality generated with GPT-4o from scratch with provided regions. Errors are highlighted and underlined in red.} 
    \label{fig:qual_gpt_sg}
\end{figure*}
Instead of training our own model for scene graph generation, we see if GPT-4 can generate a scene graph from scratch. Inspired by set-of-mark prompting~\cite{yang2023setofmark}, we include the segmentation masks and numerical IDs in the image, as well as the bounding box coordinates of the regions, and prompt the model to generate such scene graph. Figure~\ref{fig:qual_gpt_sg} shows an example. We observe that the GPT-generated scene graph contains blatant mistakes, lacking precise depth and grounding information. For instance, it inaccurately labels [4] as the stroller with child, while it actually refers to the car. The sidewalk is incorrectly labeled as being walked on by [1] and [2], whereas the street [6] is the correct object. The model also hallucinates that [7] is a phone in hand instead of region [9]. Based on this evaluation, we assert that generating scene graph directly from GPT4 is infeasible and our data curation pipeline, which constructs \dataset\ from a seed dataset with accurate human annotations, is necessary to train \model\ with accurate and high-quality scene graphs. Moreover, this shows that our model possesses the unique capability of generating dense scene graphs that state-of-the-art models like GPT-4 lack, highlighting the significance of our contribution.

\subsection{GPT-4 Edited Scene Graph}
\begin{figure}[ht!]
    \centering
    \includegraphics[width=0.7\textwidth]{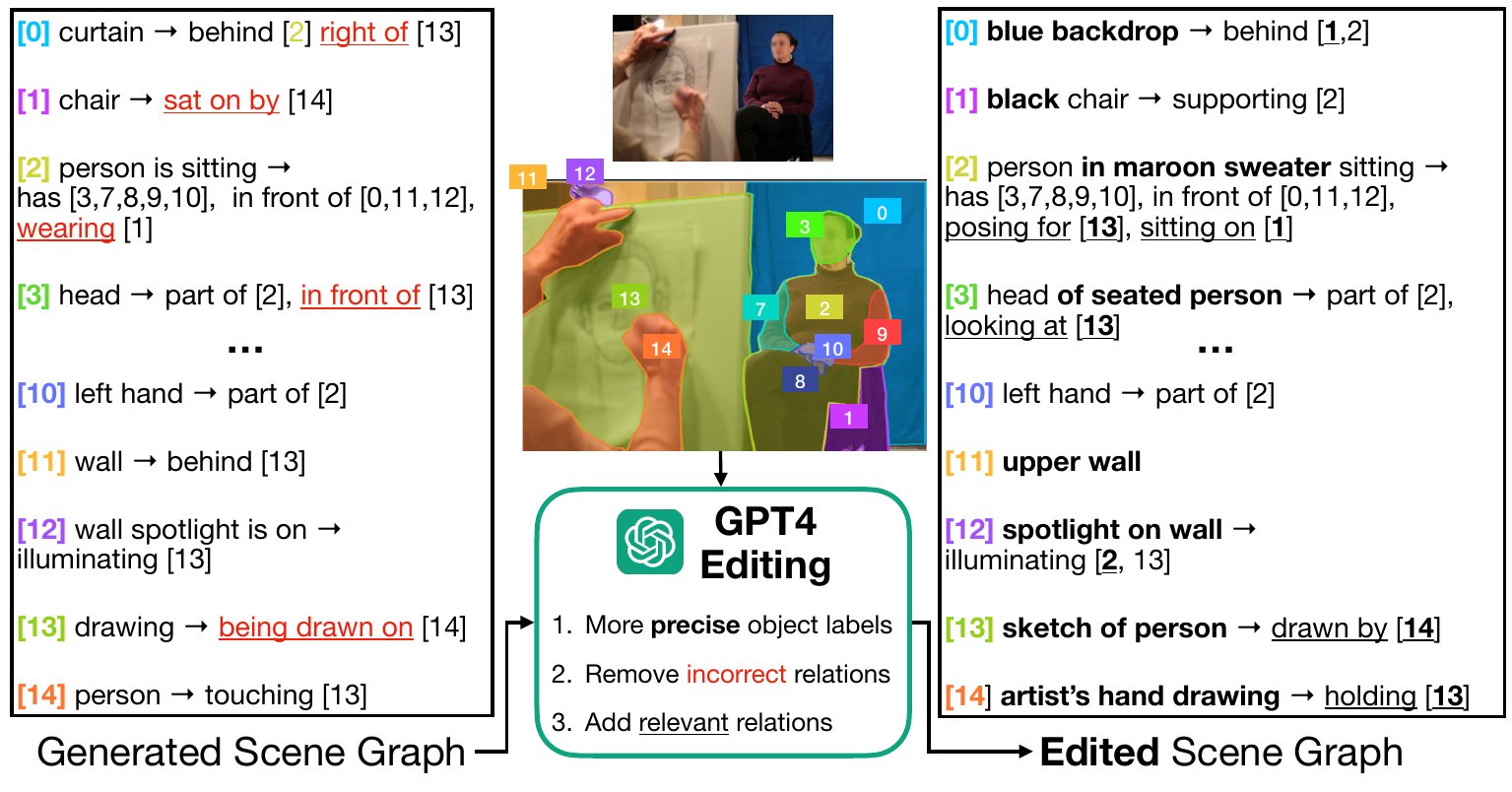}
    \caption{
    Example of \datasetsecond collection with \distill. GPT-4 can provide more precise object descriptions with attributes (bolded) and remove incorrect relations (highlighted in red) while adding relevant ones (underlined).
} 
\label{fig:sg-edit}
\end{figure}

In Figure~\ref{fig:sg-edit}, we further illustrate how the \distill~pipeline improves the quality of generated scene graphs through GPT-4o editing. We see that GPT-4 can effectively refine scene graphs by removing an irrelevant “wearing” relation with a chair, expanding the person’s description to “wearing maroon sweater,” and introducing relationships such as “posing for [13]” or “sitting on [1].”

\section{Training Details}

\subsection{Implementation details}
We train the model for two epochs for each stage. The input image is padded and resized to 512x512 single crop image. The LM has a learning rate of $1e^{-5}$ and supports max sequence length of 8192. In Stage 0, we keep the vision encoder frozen with 128 batch size and took 6 hours to train with 16 H100 GPUs. In Stage 1 and Stage 2, we train the visual encoder jointly with different learning rates of $1e^-{6}$, which is a common practice in SoTA MLMs \cite{chen2024far}. Both stages took 12 hours with 32 H100 GPUs. We use batch size of 64, linear warmup ratio of $0.03$  with cosine annealing, and Adam optimizer with DeepSpeed Stage-2 configuration \cite{deepspeed}.

\subsection{Encoding Image Regions in Pixel and Text Space}

Inspired by previous MLMs that support region-based understanding~\cite{ma2025groma, zhang2023gpt4roi, yuan2024osprey}, we represent specific regions within an image using unique numerical identifiers \texttt{region\{i\}}. These identifiers can be referenced through bounding box coordinates embedded in the text and/or via segmentation masks in the pixel space. To facilitate this, we define a special token \texttt{<region\{i\}>}, that is appended with a mask token \texttt{<mask>} and a spatial position token \texttt{<pos>}. These tokens serve as placeholders that are replaced by the corresponding region and spatial features following Osprey~\cite{yuan2024osprey}. Additionaly, we include bounding box coordinates \texttt{<|box\_start|>(x1, y1),(x2, y2)<|box\_end|>},  where $(x1, y1)$ and $(x2, y2)$ denote the normalized top-left and bottom-right coordinates of the bounding box, respectively. We normalize all bounding box coordinates to a range from 0 to 1000 to maintain consistency across different image scales. This approach allows the masked regions to be seamlessly integrated with textual content in pixel and text space, forming complete sentences within the same tokenization space. In sum, each region \texttt{region\{i\}} is represented as:  \texttt{region\{i\} <mask> <pos> <|box\_start|>(x1, y1),(x2, y2)<|box\_end|>}.



\section{Training Dataset}
\label{sec:training_dataset}
We present more details for the Stage 1 and Stage 2 datasets described in Section~\ref{sec:training_stage}. Table~\ref{tab:stage1_dataset}-\ref{tab:stage2_dataset} show the general dataset configurations. Namely, we use only \datasetfirst for Stage 1 training and only \datasetsecond for Stage 2. This results in a total of 1.73M examples used in Stage 1 training and 1.23M examples in Stage 2 training. The input instruction and output text format for grounding and scene graph tasks are additionally shown in Table~\ref{tab:prompt_grounding}-\ref{tab:prompt_dense}. For scene graph data, we use a maximum of 20 relations per object.

\begin{table*}[ht!]
\centering
\resizebox{\linewidth}{!}{%
\begin{tabular}{l|l|c|l}
\toprule
Dataset Type& Task& \# Examples & Datasets\\ \midrule
\multirow{5}{*}{Visual Instruction}& \multirow{2}{*}{VQA} & \multirow{2}{*}{430K} & VQAv2~\cite{vqav2}, GQA\cite{hudson2019gqa}, GQA-CoT~\cite{hudson2019gqa, chen2023shikra}, VSR~\cite{liu2023visual}, OKVQA~\cite{okvqa} \\
& & & ChartQA\cite{masry2022chartqa}, DocVQA~\cite{mathew2021docvqa}, DVQA~\cite{kafle2018dvqa}, TallyQA~\cite{Acharya2018TallyQAAC} \\
& Conversation & 226K & LLava-Instruct~\cite{zhang2023llava, liu2023improved}, Osprey-724k~\cite{yuan2024osprey}, LLaVAR~\cite{zhang2023llavar}\\ 
& Multiple Choice & 138K & AOKVQA~\cite{AOKVQA}, VCR~\cite{zellers2019recognition}, AI2D~\cite{kembhavi2016diagram}, ScienceQA~\cite{saikh2022scienceqa}, TQA~\cite{Kembhavi2017AreYS} \\
& Captioning & 213K & ShareGPT-4o~\cite{cai2024internlm2}, Flickr30K-Entities~\cite{plummer2016flickr30kentitiescollectingregiontophrase, chen2023shikra}, , LAION-GPT4V~\cite{zhang2023llavar}\\
& OCR & 160K & OCRVQA, SynthDoc-En, ST-VQA, TextCaps\cite{sidorov2020textcaps}\\
\midrule
\multirow{2}{*}{Grounding}& Referring Expression Comprehension & 110K & RefCOCO, RefCOCO$^+$, RefCOCOg~\cite{yu2016modeling}\\
& Region Captioning & 77K & VG Region Captions~\cite{krishna2017visual} \\
\midrule
\multirow{3}{*}{Scene Graph} & Scene Graph Detection & 118K & VG~\cite{krishna2017visual}, PSG~\cite{yang2022panoptic}\\
& Scene Graph Classification\textsuperscript{*} & 87K & VG~\cite{krishna2017visual}, PSG~\cite{yang2022panoptic}\\
& Dense Object \& Relationship Generation\textsuperscript{1} & 171K & \datasetfirst (Ours) \\

\bottomrule
\end{tabular}
}
\caption {
Dataset used in Stage1 training. See Table~\ref{tab:prompt_grounding}-\ref{tab:prompt_dense} for input and output text format used in Grounding and Scene Graph tasks. \textsuperscript{*}We merge scene graphs in VG and PSG by identifying the duplicate regions based on the IoU similarity (threshold $>0.5$) and reassign their relationships with the new set of non-duplicate regions, resulting in 87K unique images for this task. 
}
\label{tab:stage1_dataset}
\end{table*}

\begin{table*}[ht!]
\centering
\resizebox{\linewidth}{!}{%
\begin{tabular}{l|l|c|l}
\toprule
Dataset Type& Task& \# Examples & Datasets\\ \midrule
\multirow{4}{*}{Visual Instruction}& \multirow{2}{*}{VQA} & \multirow{2}{*}{430K} & VQAv2~\cite{vqav2}, GQA\cite{hudson2019gqa}, GQA-CoT~\cite{hudson2019gqa, chen2023shikra}, VSR~\cite{liu2023visual}, OKVQA~\cite{okvqa} \\
& & & ChartQA\cite{masry2022chartqa}, DocVQA~\cite{mathew2021docvqa}, DVQA~\cite{kafle2018dvqa}, TallyQA~\cite{Acharya2018TallyQAAC}, ST-VQA \\
& Conversation & 144K & Osprey-724k~\cite{yuan2024osprey} \\ 
& Multiple Choice & 83K & AOKVQA~\cite{AOKVQA}, AI2D~\cite{kembhavi2016diagram}, ScienceQA~\cite{saikh2022scienceqa}, TQA~\cite{Kembhavi2017AreYS} \\
\midrule
\multirow{2}{*}{Grounding}& Referring Expression Comprehension & 110K & RefCOCO, RefCOCO$^+$, RefCOCOg~\cite{yu2016modeling}\\
& Region Captioning & 77K & VG Region Captions~\cite{krishna2017visual} \\
\midrule
\multirow{2}{*}{Scene Graph} & Scene Graph Detection & 118K & VG~\cite{krishna2017visual}, PSG~\cite{yang2022panoptic}\\
& Dense Scene Graph Generation\textsuperscript{2} & 261K\textsuperscript{*} & \datasetsecond (Ours)\\

\bottomrule
\end{tabular}
}
\caption {
Dataset used in Stage 2 training with 1.19M instances. \textsuperscript{*}We include two variants: one includes bounding box text coordinates, while other omits them when identfying the region. 
}
\label{tab:stage2_dataset}
\end{table*}


\begin{table}[ht]
    \centering
    \begin{minipage}{\textwidth}
        \begin{subtable}[t]{\textwidth}
            \centering
            \begin{tcolorbox}[colback=gray!10, colframe=black, title=Referring Expression Comprehension]
                \begin{center}
                    \includegraphics[width=0.25\linewidth]{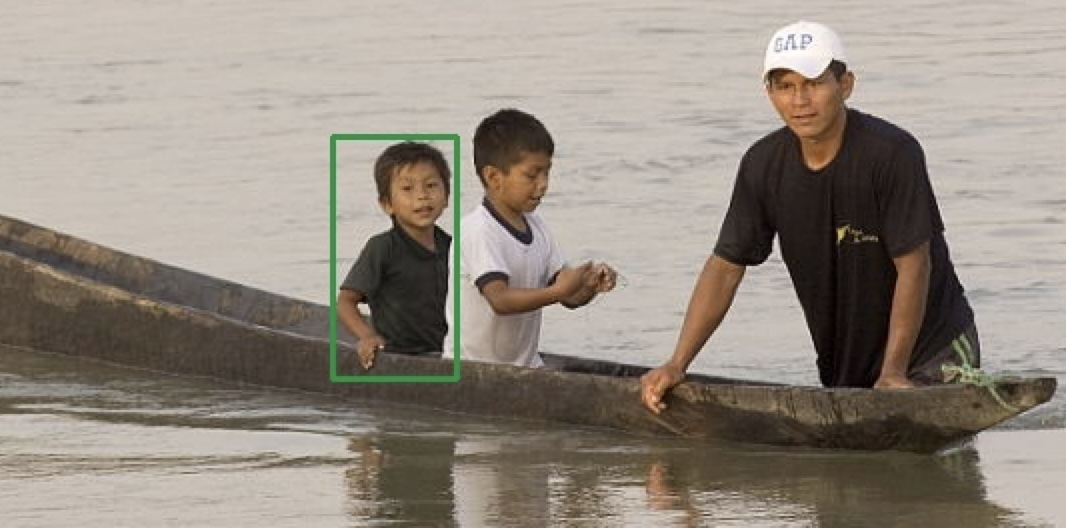}
                \end{center}
                \textbf{Input}: Please provide the region this sentence describes: small boy in black shirt.\\
                
                \textbf{Output}: \texttt{<|box\_start|>(366,515),(443,742)<|box\_end|>}  
            \end{tcolorbox}
        \end{subtable}
        \vspace{1pt}
        
        \begin{subtable}{\textwidth}        
            \begin{tcolorbox}[colback=gray!10, colframe=black, title=Region Captioning]
                \begin{center}
                    \includegraphics[width=0.25\linewidth]{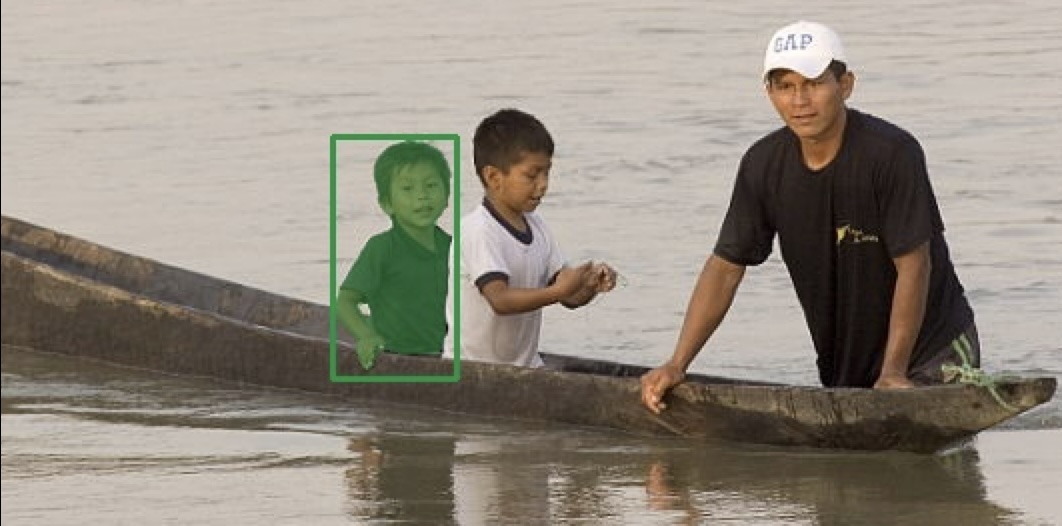}
                \end{center}
            
                \textbf{Input}: Give me a short description of: \\
                \texttt{<region> <mask><pos> <|box\_start|>(366,515),(443,742)<|box\_end|>} \\
                
                \textbf{Output}: small boy in black shirt.
            \end{tcolorbox}
        \end{subtable}
    \end{minipage}
    \caption{Example input and output for Referring Expression Comprehension and Region Captioning in Grounding task.}
    \label{tab:prompt_grounding}
\end{table}

\begin{table}[ht!]
\centering
\begin{minipage}{0.85\textwidth}
    \begin{subtable}[t]{\textwidth}
        \centering
        \begin{minipage}[t]{\textwidth}
            \begin{tcolorbox}[colback=gray!10, colframe=black, title=Scene Graph Detection]
                \begin{center}
                    \includegraphics[width=0.35\linewidth]{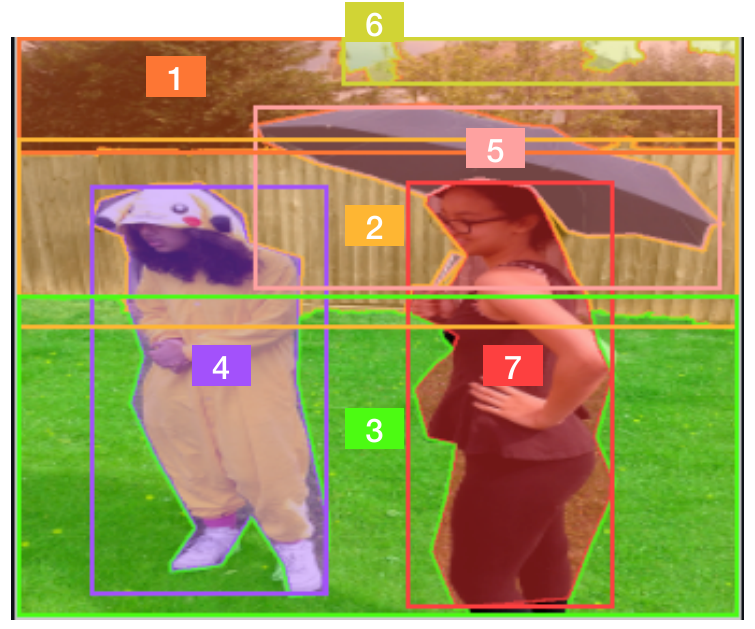}
                \end{center}
                \textbf{Input}: Generate scene graph with proposed regions in bbox coordinates.

                \textbf{Output}: \\
                Objects: \\
                \texttt{region1}: tree \texttt{<|box\_start|>(0,0),(1000,204)<|box\_end|>} \\
                \texttt{region2}: fence \texttt{<|box\_start|>(0,175),(1000,504)<|box\_end|>} \\
                \texttt{region3}: grass \texttt{<|box\_start|>(0,445),(1000,1000)<|box\_end|>} \\
                \texttt{region4}: person \texttt{<|box\_start|>(102,253),(431,964)<|box\_end|>} \\
                \texttt{region5}: umbrella \texttt{<|box\_start|>(327,117),(977,435)<|box\_end|>} \\
                \texttt{region6}: sky \texttt{<|box\_start|>(447,0),(1000,84)<|box\_end|>} \\
                \texttt{region7}: person \texttt{<|box\_start|>(539,246),(827,985)<|box\_end|>} \\
                Relations: \\
                \texttt{region2}: \texttt{region1} in front of \\
                \texttt{region4}: \texttt{region3} standing on, \texttt{region7} beside \\
                \texttt{region6}: \texttt{region1} over \\
                \texttt{region7}: \texttt{region3} standing on, \texttt{region5} holding
            \end{tcolorbox}
        \end{minipage}%
        \vspace{1pt}
    \label{tab:prompt_sg_detection}
    \end{subtable}
    \begin{subtable}[t]{\textwidth}
        \centering
        \begin{tcolorbox}[colback=gray!10, colframe=black, title=Scene Graph Classification, halign=center]
            \begin{minipage}[t]{\textwidth}
                \textbf{Input}: In the image, there are 7 regions, such as \\
                {
                \texttt{region1} \texttt{<mask> <pos> <|box\_start|>(0,0),(1000,204)<|box\_end|>}, \\
                \texttt{region2} \texttt{<mask> <pos> <|box\_start|> (0,175),(1000,504) <|box\_end|>}, \\
                \texttt{...} \\
                \texttt{region7} \texttt{<mask> <pos> <|box\_start|> (539,246),(827,985) <|box\_end|>}. \\
                }
                Generate scene graph for given regions. \\
            
            \textbf{Output}: \\
            \begin{minipage}{0.5\textwidth}
            Objects: \\
            \texttt{region1}: tree \\
            \texttt{region2}: fence \\
            \texttt{region3}: grass \\
            \texttt{region4}: person \\
            \texttt{region5}: umbrella \\
            \texttt{region6}: sky \\
            \texttt{region7}: person \\
            \end{minipage}
            \hfill
            \begin{minipage}{0.3\textwidth}
                \includegraphics[width=\linewidth]{sections/TabsNFigs/images/psg_umbrella.pdf}
            \end{minipage} \\
            Relations: \\
            \texttt{region2}: \texttt{region1} in front of \\
            \texttt{region4}: \texttt{region3} standing on, \texttt{region7} beside \\
            \texttt{region6}: \texttt{region1} over \\
            \texttt{region7}: \texttt{region3} standing on, \texttt{region5} holding
            \end{minipage}
        \end{tcolorbox}
    \label{tab:prompt_sg_classification}
    \end{subtable}
\end{minipage}
\caption{Example input and output for Scene Graph Detection and Scene Graph Classification.}
\label{tab:prompt_sg}
\end{table}

\begin{table}[t]
    \centering
    \begin{minipage}{0.8\textwidth}
        \begin{subtable}[t]{\textwidth}
        
        \begin{tcolorbox}[colback=gray!10, colframe=black, title=Dense Object \& Relationship Generation]

        \begin{center}
            \includegraphics[width=0.25\linewidth]{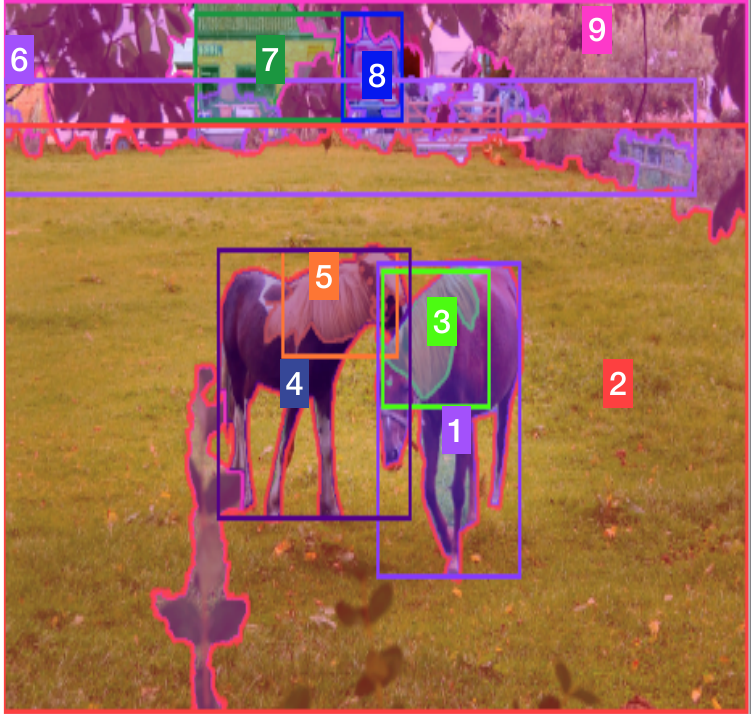}
        \end{center}
        
        \textbf{Input}: In the image, there are 9 regions, such as: \\
        \texttt{region1 <mask><pos> <|box\_start|>(0,172),(1000,1000)<|box\_end|>}, \\
        \texttt{region2 <mask><pos> <|box\_start|>(579,73),(617,113)<|box\_end|>}, \\
        \texttt{...} \\ 
        \texttt{region9 <mask><pos> <|box\_start|>(0,0),(1000,1000)<|box\_end|>}. \\ 
        Generate a description for: \texttt{region4}. \\
        
        \textbf{Output}: A dark brown horse with white spots and a mane.
        
        \vspace{6pt}
        \hrule
        \vspace{6pt}
        
        \textbf{Input}: Generate a list of relationships for: \texttt{region4}. \\
        
        \textbf{Output}: \texttt{region5} has, \texttt{region2} stands on top of, \texttt{region1} nudging, \texttt{region1} left of, \texttt{region6} in front of
        \end{tcolorbox}
        
        \begin{tcolorbox}[colback=gray!10, colframe=black, title=Dense Scene Graph Generation]
        \begin{center}
            \includegraphics[width=0.25\linewidth]{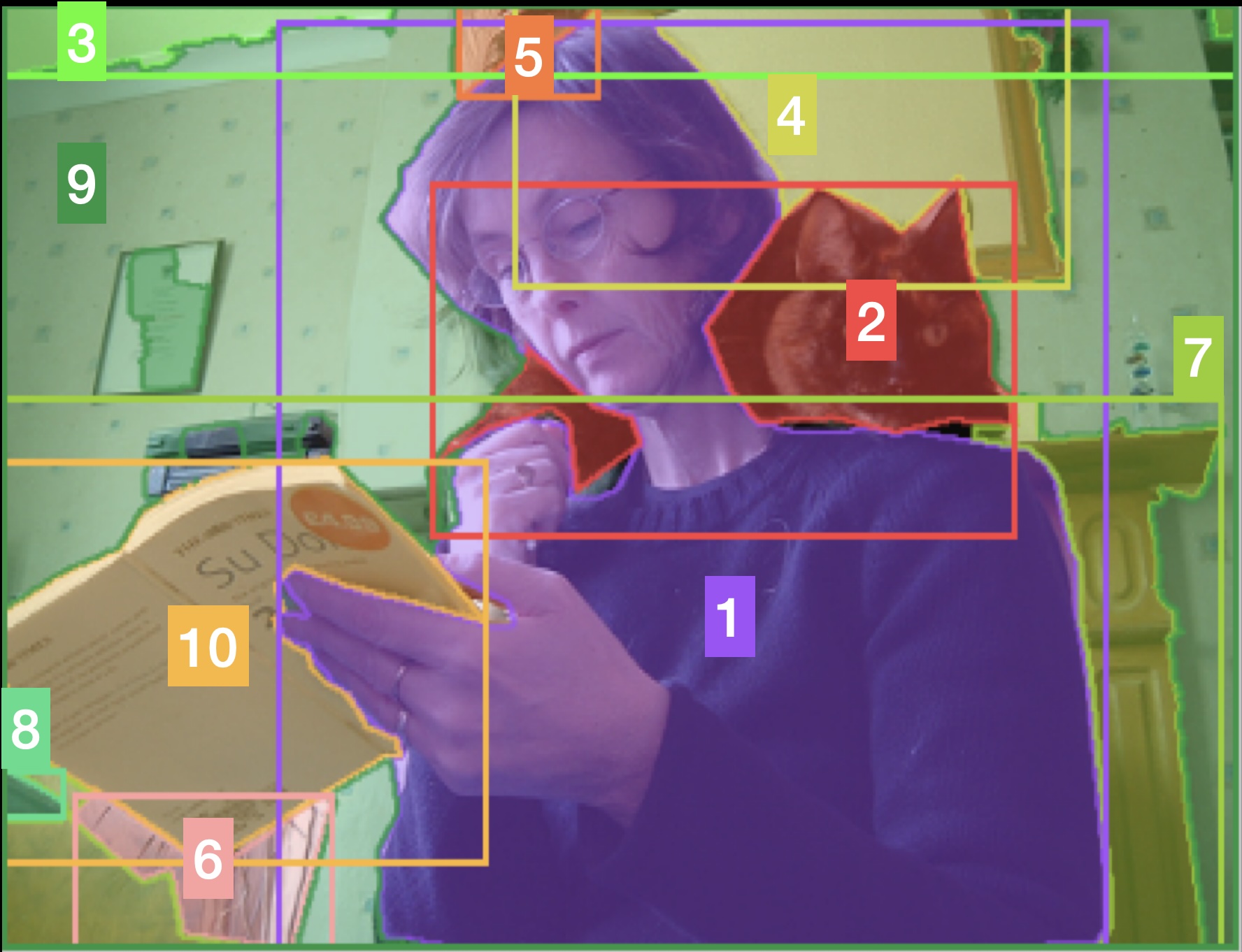}
        \end{center}
        \textbf{Input}: In the image, there are 10 regions, such as: \\ 
        \texttt{region1 <mask><pos> <|box\_start|>(222,19),(894,1000)<|box\_end|>}, \\
        \texttt{region2 <mask><pos> <|box\_start|>(346,187),(820,566)<|box\_end|>}, \\
        \texttt{...} \\
        \texttt{region10 <mask><pos> <|box\_start|>(0,480),(392,912)<|box\_end|>}. \\
        Generate an extremely detailed scene graph for given regions. \\

        \textbf{Output}: \\
        Objects: \\
            \texttt{region1}: person wearing a black sweater, reading a book, and holding a cat. \\
            \texttt{region2}: cat sitting on woman's shoulder. \\
            \texttt{...} \\
            \texttt{region10}: book being read by a woman. \\
        Relations: \\
        \texttt{region1}: \texttt{region2} caring for, \texttt{region4} in front of, \texttt{region9} in front of,  \texttt{region10} holding, \texttt{region10} engaged with, \texttt{region10} reading, \texttt{region10} looking at \\
        \texttt{region2}: \texttt{region1} being comforted by, \texttt{region1} attached to,
         \texttt{region1} sitting on \\
         \texttt{...} \\
        \texttt{region10}: \texttt{region1} read by, \texttt{region1} held by, \texttt{region7} in front of
        
        \end{tcolorbox}
    \end{subtable}
\end{minipage}
\caption{Example input and output for Dense Object \& Relationship Generation and  Dense Scene Graph Generation.}
\label{tab:prompt_dense}
\end{table}


\section{Evaluation Details}
\label{sec:eval_details}
This section provides detailed information on the evaluation configurations used for the visual reasoning tasks presented in our experiments. Unless explicitly stated otherwise, all models are evaluated using greedy decoding (i.e., selecting the token with the highest probability at each step) without sampling. The sole exception is the scene graph detection task, where we employ a temperature of $0.2$ and a top-$p$ value of $1.0$ during generation \cite{holtzman2020curious} .

\subsection{Relationship Understanding}

We evaluate relationship understanding using various benchmarks, each with specific evaluation protocols:

\paragraph{GQA~\cite{hudson2019gqa}:} The model's performance is assessed by the accuracy of exact matches between its answers and the ground truth answers. The prompt provided to the model is as follows: \textit{Question: \{\texttt{question}\} Answer the question using a single word or phrase.}

\paragraph{Visual-Semantic Reasoning (VSR):} We frame the VSR task as a binary classification problem by asking the model yes/no questions if the mentioned subject, relation, and object triplet holds true in the image. The model's responses are evaluated for correctness in indicating true or false statements. The prompt is: \textit{Question: Is the \{\texttt{subj}\} \{\texttt{relation}\} \{\texttt{obj}\}? Answer with yes or no.}

\paragraph{Other Benchmarks (MMBench, SeedBench-Image, CRPE, What's Up):} For these multiple-choice tasks, we prompt the model to select the correct option from a set of lettered choices. The model's accuracy is measured based on its selection of the correct option. The prompt used for multiple choice benchmarks is: \textit{Question: \{\texttt{question}\} Choices: \{\texttt{choices}\} Answer with the option's letter from the given choices directly.}

\subsection{Referring Expression Comprehension}

For the referring expression comprehension task, the model is prompted to identify the region described by a given sentence. The prompt is: \textit{Please provide the region this sentence describes: \{\texttt{sentence}\}}

The model's response is expected to contain bounding box coordinates corresponding to the described region. We extract these coordinates and compare them against ground truth annotations for evaluation.

\subsection{Region Classification}

In the region classification task, we aim to determine the category of a given region in the image. The model is prompted as follows: \textit{What is the category of \texttt{<mask><pos> <box>}? Answer using only one word or phrase.}

Here, \texttt{<mask><pos>} represents the segmentation mask and positional information, and \texttt{<box>} is the bounding box coordinates. Unlike Osprey~\cite{yuan2024osprey}, which uses only the segmentation mask, we incorporate bounding box information inferred from the mask to enable more precise object localization.

To evaluate the model's predictions, we use the Sentence-BERT\footnote{\texttt{sentence-transformers/all-MiniLM-L6-v2}} model~\cite{reimers2019sentence} to compute the semantic similarity between the predicted category and the ground truth class. We encode both the predicted and ground truth class names and compute the cosine similarity between their embeddings.

\subsection{Panoptic Segmentation}

For the panoptic segmentation task, we employ the ADE20K dataset~\cite{zhou2019semantic}. The model is prompted with: \textit{What is in \texttt{<mask><pos>}? Answer using a short phrase.}

After obtaining the model's response, we map the predicted description to the closest object class using semantic similarity, as in the region classification task, utilizing the same Sentence-BERT model. We encode the object class names and compute the cosine similarity between the embeddings of the predicted and ground truth classes. Standard segmentation metrics, including those for panoptic, instance, and semantic segmentation, are used to evaluate the model's performance\footnote{\texttt{https://github.com/facebookresearch/detectron2/blob/main/detectron2/evaluation/panoptic\_evaluation.py}}.

\subsection{Scene Graph Detection}

In the scene graph detection task, the model is prompted to generate a scene graph, including object regions and their relationships. The prompt is: \textit{Generate a scene graph with proposed regions in bounding box coordinates.}

An illustrative example of the expected input and output format is provided in Table~\ref{tab:prompt_sg}. We extract the bounding box regions, object descriptions, and relationship triplets from the model's output using regular expressions. To assign labels in the Panoptic Scene Graph (PSG) dataset, we match the object descriptions to the closest object classes by computing the maximum cosine similarity between their embeddings, as in the panoptic segmentation task. Specifically, we encode phrases of the form ``The object is \{\texttt{class\_name}\}'' for both the ground truth and predicted object descriptions.

For determining the best predicate (relationship), we encode the relationships using the phrase ``Object is \{\texttt{predicate}\} another object'' and compute the cosine similarity between the embeddings of the predicted and ground truth relations.

Following the recall-based evaluation protocol in scene graph literature~\cite{krishna2017visual}, a predicted triplet \texttt{(subject, predicate, object)} is considered correct if:

\begin{enumerate}
    \item The predicted subject and object bounding boxes have an Intersection over Union (IoU) greater than 0.5 with the ground truth bounding boxes.
    \item The predicted subject class, predicate, and object class match those of the ground truth.
\end{enumerate}

\section{\dataset Data Generation Details}
\label{sec:svg_relations_appendix}
In this section, we provide more details of the data generation pipeline.

\subsection{\textbf{\datasetfirst}}
\begin{figure*}[t]
    \centering
    \includegraphics[width=0.9\textwidth]{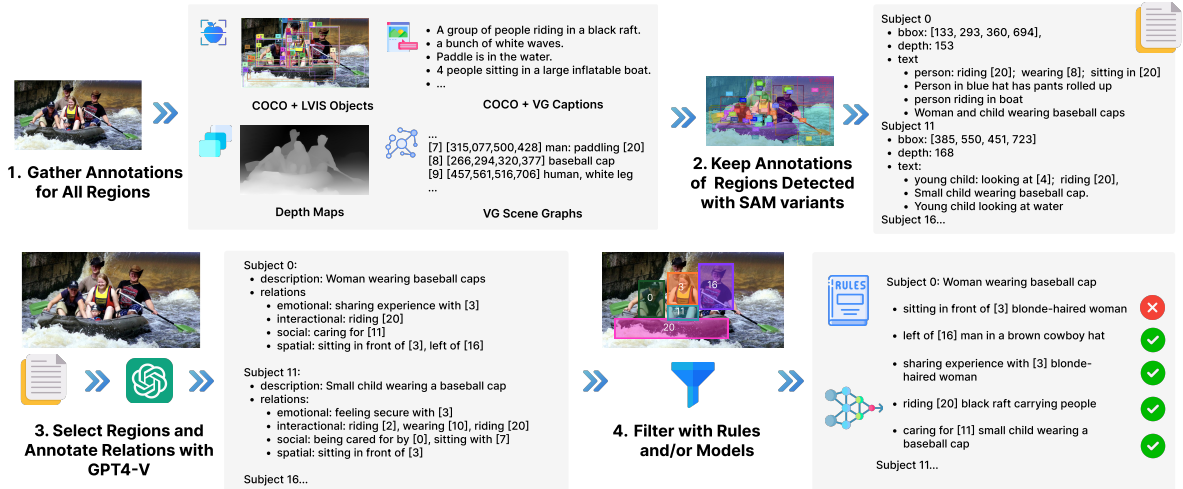}
    \caption{
    \small
    Data generation pipeline for \datasetfirst, consisting of four steps: (1) gathering annotations for all the regions; (2) keeping only the annotations of the regions detected by SAM-SEEM; (3) selecting and annotating regions with GPT4-V and (4) filtering out incorrect relations with rules and/or vision-language models such as CLIP or LLaVA.
    \label{fig:data-gen}
    }
\end{figure*}

Figure ~\ref{fig:data-gen} shows an overview of data curation for \datasetfirst in four steps. We describe the last three steps in more detail.

\paragraph{Selecting semantically significant objects} We filter out object regions of low semantic significance to retain visually distinct and meaningful elements. Specifically, we use Segment Anything (SAM)~\cite{kirillov2023segment} and Semantic-SAM~\cite{li2023semantic} to generate segmentation masks representing prominent objects and regions in each image. For each annotated region in our seed dataset, we compute the Intersection over Union (IoU) score between our annotated regions and the segmentation masks produced by SAM and Semantic-SAM. We then keep annotated regions with an IoU score greater than 0.5 with any of the segmentation masks. We discuss the configurations region proposal and their refinement in Appendix~\ref{sec:proposals}.

\paragraph{Generating relationships with GPT-4V}
Building on the dense region annotations, we prompt GPT4-V to identify at lest $K=5$ subjects in each scene. For each subject, we request 1) its description, and 2) a comprehensive list of its relationships with other objects, categorized into five types: spatial, interactional, functional, social, and emotional. The prompt for calling GPT-4V is shown in Table~\ref{tab:prompt_svg}.

\paragraph{Filtering irrelevant relationships}\label{sec:filtering}
Despite using human annotations for object regions, GPT4-V is prone to generating relationships with errors. 
Thus, we implement a filtering strategy to remove such inconsistencies ~\cite{fang2023data, Park2023LocalizedSK}.  
Specifically, we apply rule-based filtering to spatial relationships and model-based VQA filtering to the rest. In Appendix~\ref{sec:dataset_filtering}, we elaborate on these filters and their impact on relationship distribution.

\begin{table}[t!]
\begin{pastelbox}[pastelgreen]{Prompt for Dense Object \& Relationship Generation}
\textbf{System Prompt}: You are provided 'Captions', 'Region Captions', 'Scene Graph', and 'QAs' annotated by humans, which are guaranteed to be correct. \\
- [Captions]: General image overviews. \\
- [Region Captions]: Detailed descriptions of particular image regions. \\
- [Objects]:  Grounded objects annotated by humans. In format of [object\_id (int)]: [bbox] object\_name \\
- [Scene Graph]: In format of [object\_id (int)]: ([bbox], optional depth) [list of attributes] object\_name [relationship with other objects] \\
Bounding boxes (bbox) are labeled in the form of bounding box as [x1,y1,x2,y2] normalized from 0 to 1000.
You might also have depth information from 0 to 255, where low depth means it's farther back in the image.

**Objective**: Your task is to meticulously document all discernible relationships between a selected object and other objects within the scene, and provide a dense, detailed description of the object as well. This involves a comprehensive examination of the scene to identify and record every possible, interesting relationship that the chosen object has/ may have with others. These relationships can be of various types, including but not limited to interactional, (e.g., holding, looking at), functional (e.g., part of, used by, owned by, contains), hierarchical, and spatial relationships (e.g., above, behind, attached to, etc.). \\ Please specify the category of relationship you are looking to annotate and then provide the list. \\

**Relation Requirement**:
- Try to include at least 10 total relationships per object, describing beyond what is said explicitly in the descriptions if needed, by reasoning about the bounding boxes and depth.  \\
- Prioritize including interactional and action-based relationships first. \\
- Prioritize human-annotated captions for accuracy and consistency, and carefully consider model-generated descriptions for additional details. \\
- Categorize your relationships into one of the following categories: \\
    1. **Spatial Relationships**: These refer to the physical location or position of the person in relation to other objects or individuals. Examples include 'above', 'below', 'over', 'across from', 'behind', 'in front of', 'inside', 'outside' etc. \\

    2. **Interactional Relationships**: These involve some form of action or interaction between the person and other objects or individuals. Examples include 'holding', 'touching', 'looking at', 'talking to', 'playing with', 'using', etc. \\

    3. **Functional Relationships**: These refer to the purpose or function of an object in relation to the person. Examples include 'worn by', 'used by', 'owned by', 'part of', etc. \\

    4. **Social Relationships** (for human): These refer to the social connection or interaction between the person and other individuals. Examples include 'friend of', 'sibling of', 'parent of', 'colleague of', 'boss of', etc. \\

    5. **Emotional Relationships** (for human): These refer to the emotional connection or feelings between the person and other individuals. Examples include 'loves', 'likes', 'dislikes', 'hates', etc. \\

**Description Requirement**:
- The description should cover the interesting features and focus on relationships with other objects.
- Your description should be factual and to the point. It SHOULD NOT have suggesting, speculative or interpretative additions about the object's significance or its contribution to the scene's atmosphere and vibe.
- **Physical Attributes**: Briefly detail the object's color, size, texture, attributes, and any other pertinent physical characteristics.
- **Position**: Describe the object's location within the scene with precision.
- **Relationships with Other Objects**: Document all direct spatial and interactional relationships the object has with other objects. This includes being next to, above, below, holding, or any other clear relationship. Each relationship should be stated plainly, specifying the type of relationship and the related object's ID.
- Avoid including obvious traits/properties of the object that can be inferred without looking at the image. \\

\end{pastelbox}
\caption{Prompt for collecting \datasetfirst.}
\label{tab:prompt_svg}
\end{table}
\begin{table}[t]
\begin{pastelbox}[pastelblue]{Prompt for Scene Graph Edit}
\textbf{System Prompt}: You are given the original image and a set of same images with highlighted regions specified by unique id labels (colors matching the referred region). 
You are provided: \\
    - dense caption of the image that describes the scene as detailed as possible. \\
    - scene graph **generated** by AI model that might contain hallucinated objects and relationships. \\
    
Your goal is to fix and improve this AI generated scene graph.
Each object in the scene graph is labeled with a unique numeric id and provided with bounding box coordinates `[x1, y1, x2, y2]` (top-left and bottom-right).
We want to verify the correctness of the objects and relationships in the scene graph.
The input scene graph has the format: \\
{
    'obj\_id': {
        'name': 'A description of the object',
        'bbox': [x1, y1, x2, y2],
        'rel': {
            'rel\_name': [integer list of other obj\_ids],
        }
    },
}
\\

Your task: 
Generate only the ''Edits'' needed to correct the objects and relationships in the scene graph.
Your response will be called by python dictionary `update` method, so make sure you only generate just the NECESSARY edits. \\

1. Go through all the objects in the scene graph and determine if you can confidently say the object mentioned by noisy description is really present and visible in the image.  \\
    - If you think you can confidently verify the presence of an object and the description can be improved, provide a revised description with more accurate object name and more visual specificity. \\
    - If you are unable to confidently verify the presence of an object or suspect it might be inaccurately represented for the given region, 
    please provide a revised description using a more accurate object name if possible, or alternatively, classify it under a broader, safer category. 
    These can be 'decoration', 'tool', 'foliage', 'furniture item', etc., to ensure clarity and avoid misidentification.
    Then, provide more specificity to your name so that viewer can disambiguate the object from other objects. \\
    - Try to include the object's position, color, texture, and/or any other distinguishing feature. \\
    - Remember to use the first original image to accurately identify the color and texture of the objects.  \\
    - Make sure to take the bounding box size and location into account to correctly describe the specified region. \\
    
2. Then, make edits to the relationships by adding the prominent relationships current scene graph is missing,
    or/and removing erroneous relationships between objects in the scene. 
    For example, object 1 has relationship: {{'on': [2,3,4,5]}} and you think it should be {{'on': [2,3]}},
    then you should include object\_ids [4,5] in the 'remove' list.
    Also, if object 2 has relationship: {{'contains': [2,6,7]}} and you think it should be {{'contains': [2,6,7,8]}},
    then you should include object\_ids [8] in the 'add' list, and also add the relationship type 'spatial' to the 'add' list.
    If no relationships need to be added and/or removed, please return an empty list for 'add' and/or 'remove'.

Output should be a flattened JSON object with no unnecessary spacing.
\end{pastelbox}
\caption{Prompt to create \datasetsecond using \distill }
\label{tab:prompt_svg_edit}
\end{table}

\subsection{Data Filtering Details} 
\label{sec:dataset_filtering}

\begin{table}[ht!]
\normalsize
\centering
\resizebox{\linewidth}{!}{%
\begin{tabular}{llccccc}
\toprule
& Version  & Images  & Relations & Objects & Relations per region / subject & Spatial / functional / interaction / social / emotional relations\\ \midrule
Visual Genome~\cite{krishna2017visual}   & COCO images      & 38K                  & 843,576               & 957,951                  & 0.88 / 1.53                              & ---                  \\
\midrule
\multirow{2}{*}{SVG-Relations} & Raw      & \multirow{2}{*}{33K} & 1,457,849               & 626,516                  & 2.33 / 6.50                              & 702,916 / 189,262 / 500,240 / 34244 / 31,187                                  \\
& Filtered &                        & 855,573                & 445,131                  & 1.92 / 4.98                              & 460,098 / 81,789 / 277,029 / 18,496 / 18,161      \\
\bottomrule
                
\end{tabular}}
\caption {
Effect of filtering pipeline while curating \datasetfirst dataset.
}
\label{tab:dataset_filtering_stats}
\end{table}

\begin{table}[ht!]
\centering
    \begin{tabular}{lcccc}
    \toprule
    Dataset & VSR & CRPE & SugarCrepe & What's Up ?\\
     & ZS-test & Relation & Relation & Controlled \\
    \midrule
    SVG without Filtering & 67.8 & 59.8 & 86.8 & 67.0 \\
    SVG with Filtering & 69.7 & 64.8 & 87.9 & 76.7\\
    \bottomrule
    \end{tabular}
    \caption{Results on relationship reasoning benchmarks comparing the Stage 1 model trained with and without filtering of the SVG dataset. We keep the same set of instruction tuning training data in Stage 1 but only change the Dense Object \& Relationship task.}
    \label{tab:ablation_filtering}
\end{table}
We show impacts of data filtering described in \datasetfirst. Table~\ref{tab:ablation_filtering} shows the relationship reasoning performance of the Stage 1 model trained with and without filtering. We see that filtering is crucial for achieving high performance and can lead to significant drop in relationship reasoning without the module. We next describe our filtering approach used to curate the data.

\paragraph{Rule-based filtering}
\begin{table*}[!ht]

\centering
\includegraphics[width=1.1\textwidth]{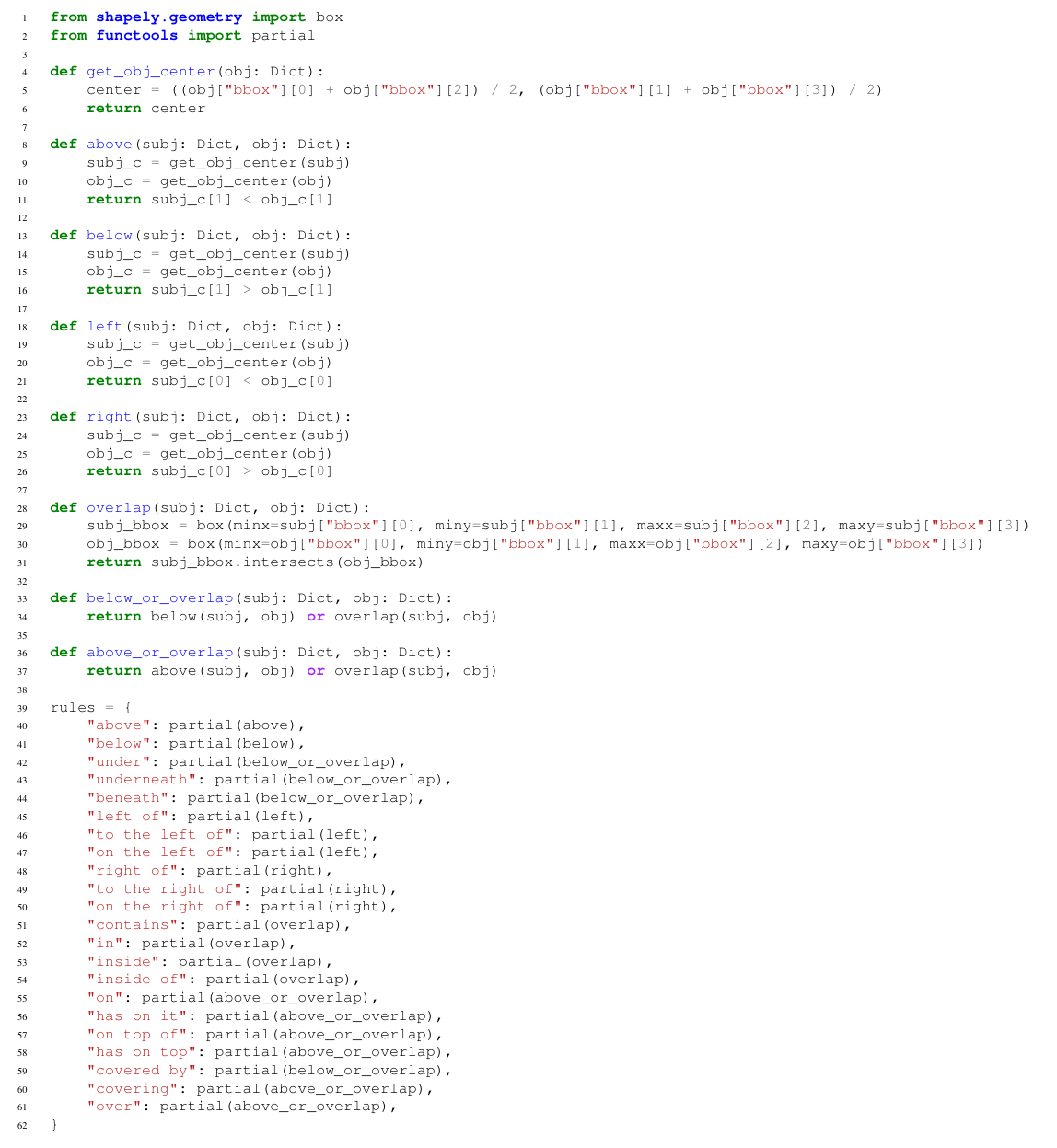}
\label{rule-based-program}
\caption{Definitions of filtering rules for spatial relations.}

\end{table*}

We implement seven simple rules -- above, below, left, right, overlap, above or overlap, and below or overlap -- based on naive physics knowledge. 
These rules cover 22 frequent spatial relationships, which represent 42.3\% of all generated relationships.
We present the definitions of all 7 rules as well as the rules used for the 22 frequent relation phrases in Listing 1. We verified the validity of these rules by getting the rule-based scores of the 102,233 Visual Genome examples containing the 22 relation phrases. Assuming Visual Genome's human annotations are correct, we treat all examples as positive. Our rule-based program predicts 1 for 95.1\% of these examples, meaning that its accuracy is 95.1\%.

\paragraph{Model-based filtering}
For relations that cannot be easily verified through rules such as functional and emotional relationships, we rely on model-based filtering using multi-modal models.

We first test out existing MLMs and formulate the verification as a binary VQA task, asking ``Does this relation correctly describe the image? Answer with Yes or No.''. We experiment with three state-of-the-art multi-modal models -- LLaVA, QwenVL, and CogVLM \cite{liu2024visual, bai2023qwen, wang2023cogvlm} -- and different combinations of their answers. Next, we try filtering with CLIP based models, and calculate the similarity scores between each ``subject relation object'' phrase and the region crop containing the subject and object and filter out examples with similarity scores lower than 0.3 \cite{schuhmann2021laion}. 

We compare these models' performance on (1) existing image-text evaluation datasets that require relation understanding such as SugarCREPE \cite{hsieh2024sugarcrepe} and CREPE \cite{ma2023crepe} and (2) a small subset (N = 600) of the data with one author's annotations.
We present these results in Tables \ref{tab:existing_eval} and \ref{tab:human_eval} respectively. Through these experiments, we find it most optimal to use both LLaVa-v1.6-vicuna-13b and Qwen-VL-Chat and filter out relationships where either model answers ``No''.  Meanwhile, we find CLIP-based filtering ineffective for most relationship types, and not use their filtering scheme.

\begin{table*}[htb]
\normalsize
\centering
\scalebox{.95}{
\begin{tabular}{lllll}
Dataset                                        & Total \# of examples    & Model        & Balanced acc    & Precision       \\ \toprule
\multirow{6}{*}{SugarCREPE (replace relation)} & \multirow{6}{*}{6,354}  & Best CLIP    & 0.7475          & —               \\
                                               &                         & LLaVa        & 0.7224          & 0.6583          \\
                                               &                         & Qwen         & 0.7398          & 0.6703          \\
                                               &                         & CogVLM       & 0.6139          & 0.5652          \\
                                               &                         & LLaVa + Qwen & \textbf{0.7581} & \textbf{0.7034} \\
                                               &                         & Majority     & 0.7181          & 0.6444          \\ \midrule
\multirow{5}{*}{CREPE (systematicity-laion)}   & \multirow{5}{*}{47,664} & LLaVa        & 0.7698          & 0.7335          \\
                                               &                         & Qwen         & 0.7605          & 0.6955          \\
                                               &                         & CogVLM       & 0.5377          & 0.5228          \\
                                               &                         & LLaVa + Qwen & \textbf{0.7954} & \textbf{0.7788} \\
                                               &                         & Majority     & 0.7361          & 0.6672     \\ \bottomrule    
\end{tabular}
}
\caption {We report the balanced accuracy and precision of different model-based filtering methods on existing benchmarks -- SugarCREPE and CREPE -- that require relation understanding.} 
\label{tab:existing_eval}
\end{table*}
\begin{table*}[]
\normalsize
\centering
\scalebox{1}{
\begin{tabular}{lllll}
\toprule
Relation types       & Total \# of examples & Method                           & Balanced accuracy & Precision       \\ \midrule
\multirow{6}{*}{All} & \multirow{6}{*}{600} & CLIP similarity \textgreater 0.3 & 0.5008            & 0.7089          \\
                     &                      & LLaVa VQA                        & 0.6163            & 0.8099          \\
                     &                      & Qwen VQA                         & 0.5486            & 0.7319          \\
                     &                      & CogVLM VQA                       & 0.5029            & 0.7095          \\
                     &                      & LLaVa + Qwen                     & \textbf{0.6165}   & \textbf{0.8165} \\
                     &                      & Majority VQA                     & 0.5484            & 0.7311  \\ \bottomrule       
\end{tabular}
}
\caption {We report the balanced accuracy and precision of different model-based filtering methods on the human-annotated set of 600 randomly sampled examples across all five relation types.} 
\label{tab:human_eval}
\end{table*}

\paragraph{Relationship distribution before and after filtering}
\begin{figure*}[ht!]
    \centering
    \includegraphics[width=0.8\textwidth]{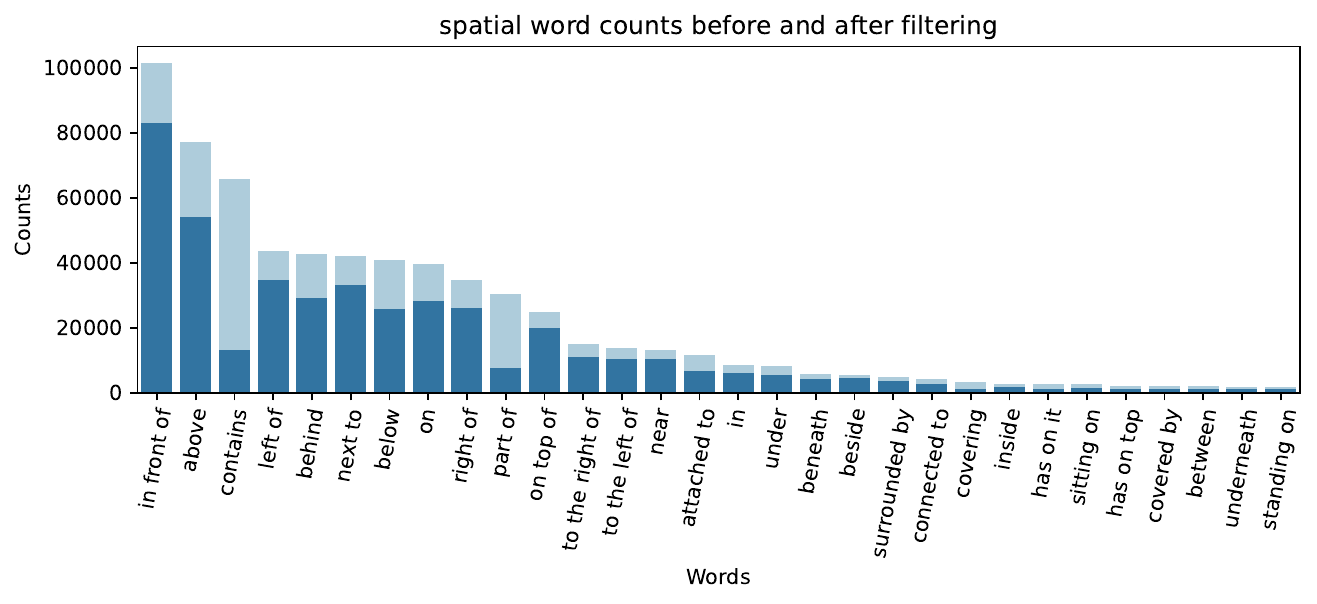}
    \caption{Distribution of spatial relations before and after filtering} 
    \label{fig:spatial_dist}
\end{figure*} 
\begin{figure*}[ht!]
    \centering
    \includegraphics[width=0.8\textwidth]{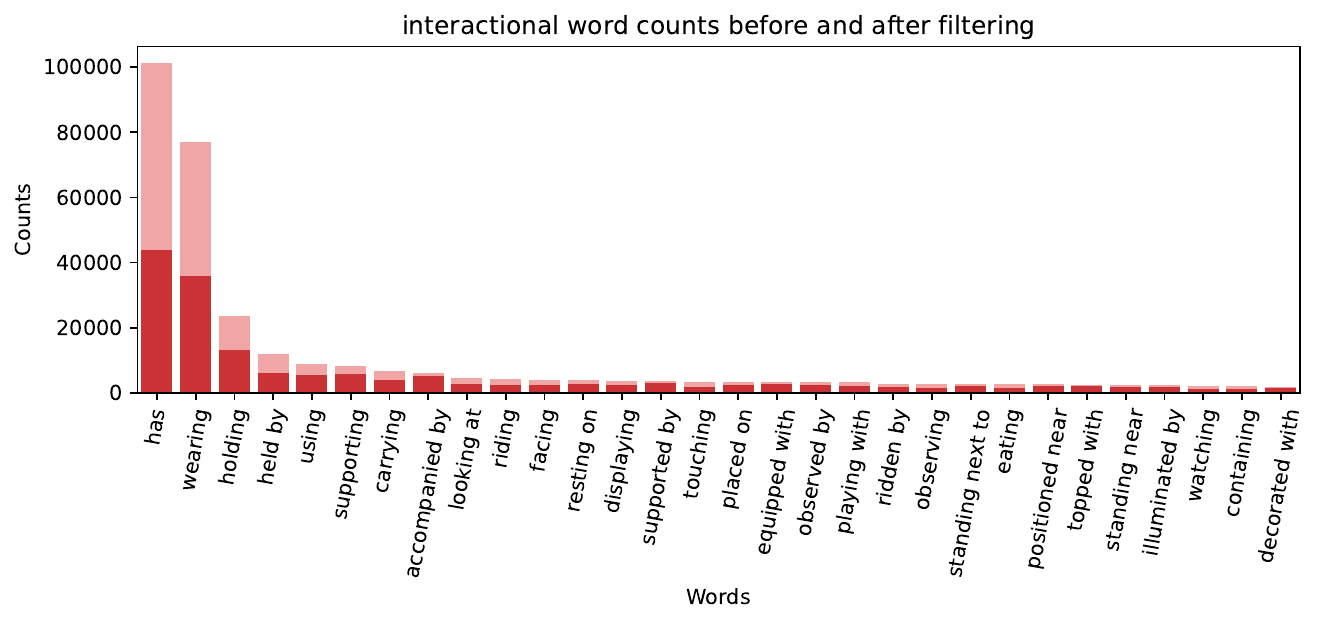}
    \caption{Distribution of interactional relations before and after filtering} 
    \label{fig:interactional_dist}
\end{figure*} 
\begin{figure*}[ht!]
    \centering
    \includegraphics[width=0.8\textwidth]{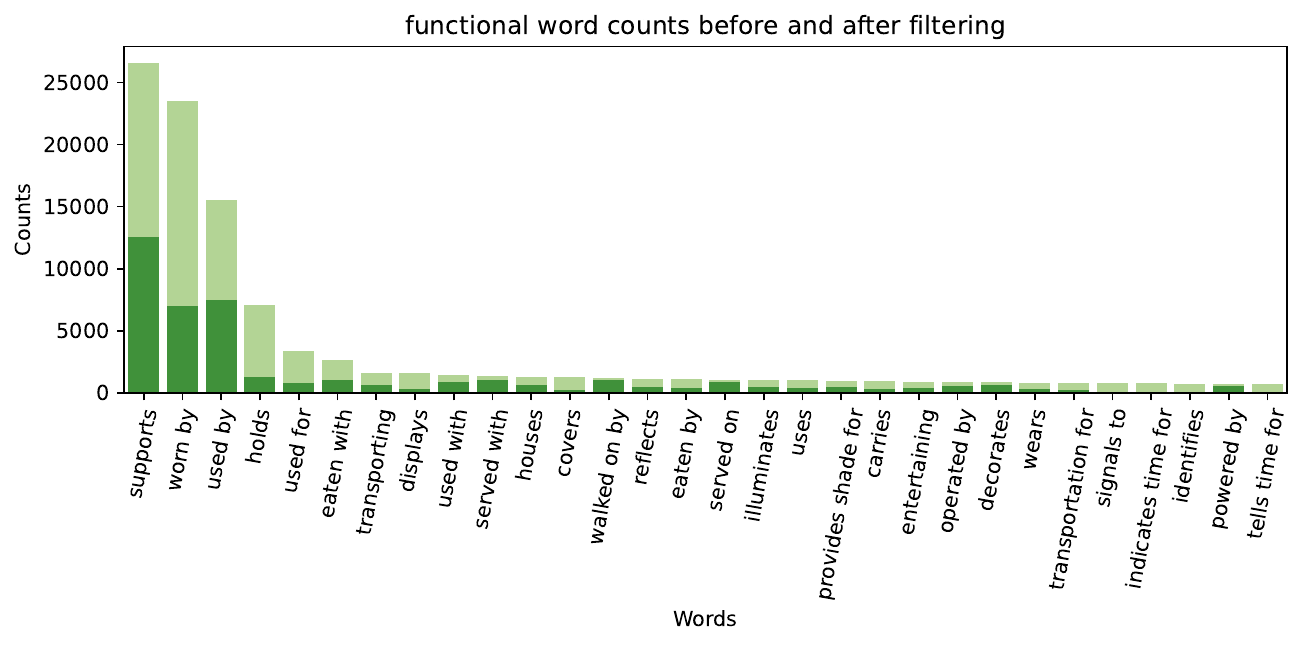}
    \caption{Distribution of functional relations before and after filtering} 
    \label{fig:functional_dist}
\end{figure*} 
\begin{figure*}[ht!]
    \centering
    \includegraphics[width=0.8\textwidth]{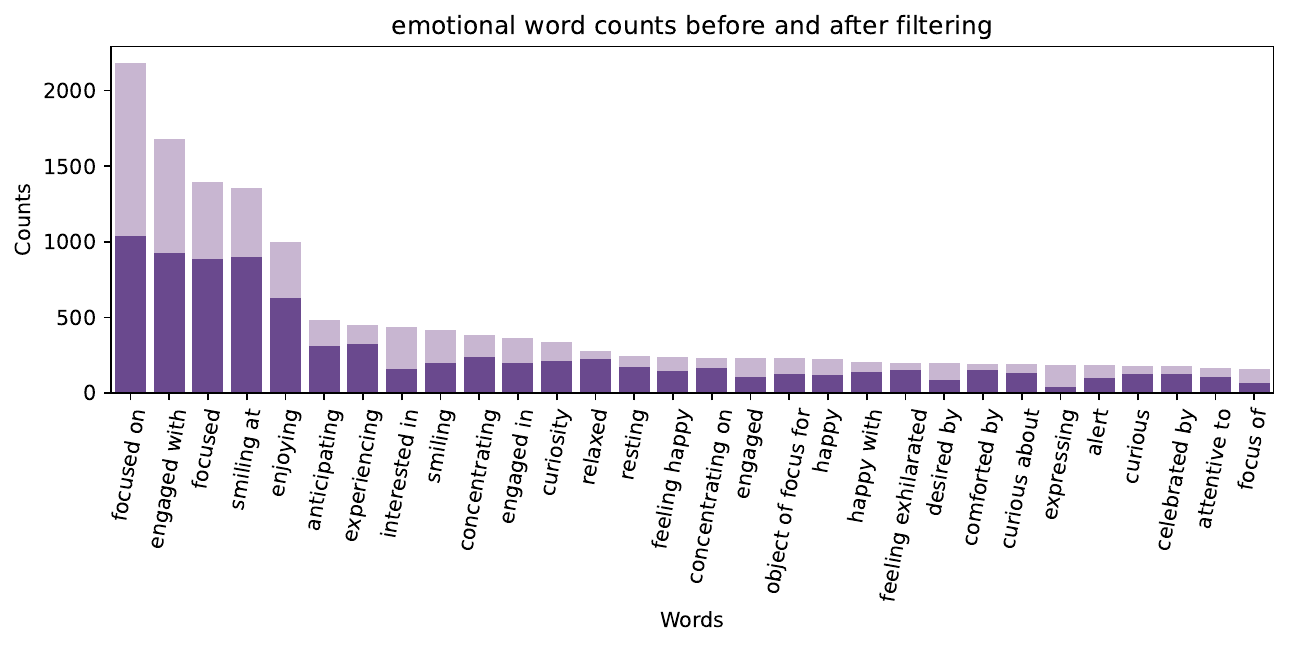}
    \caption{Distribution of emotional relations before and after filtering} 
    \label{fig:emotional_dist}
\end{figure*} 
\begin{figure*}[ht!]
    \centering
    \includegraphics[width=0.8\textwidth]{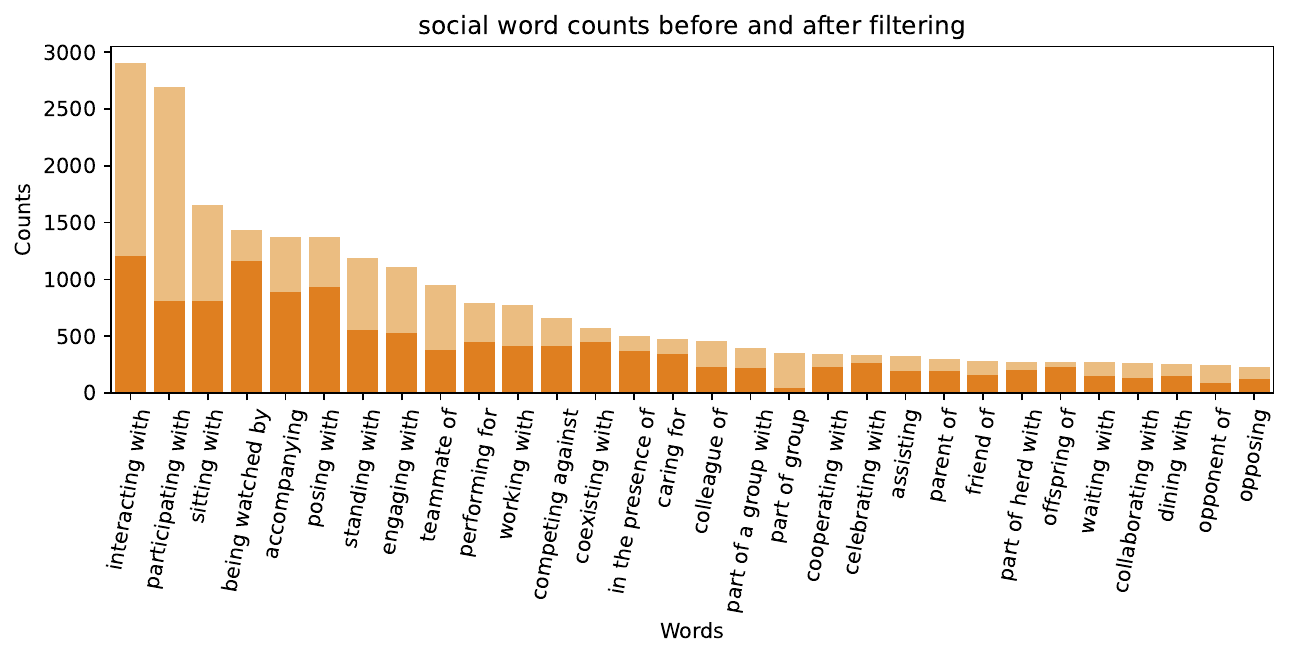}
    \caption{Distribution of social relations before and after filtering} 
    \label{fig:social_dist}
\end{figure*} 
We present the distributions of the most frequent relationships for each type of relationships before and after running our filtering pipeline in Figures \ref{fig:spatial_dist} - \ref{fig:social_dist}.


\subsection{Object Proposal Generations}\label{sec:proposals}
\begin{itemize}
    \item SAM~\cite{kirillov2023segment}: We explore the subpart, part, and whole object modes available in the SAM model as region candidates for scene graph generation. We found the part and whole mode to be relevant for covering  interesting objects in the scene. 
    \item Semantic-SAM~\cite{li2023semantic}: To extract more semantic-aware object proposals, we also employ Semantic-SAM at different levels of granularity. Specifically, 3 different granularity prompts were used, (1) granularity prompt 1, (2) granularity prompt 2 and (3) all 6 granularity prompts, as introduced in~\cite{li2023semantic}, where granularity prompts 1, 2 correspond to different semantic-level object proposals, while all 6 granularity prompts give more fine-grained, part-like region proposals. 
    \item Proposal Refinement via Union Strategy: We take the union of aforementioned two SAM and three Semantic-Sam configurations to get the final candidates. We perform non-maximum suppression with IoU threshold of $0.6$. We then sort the objects based on their area instead of stability and detection scores and take the $K$ largest objects, which captures objects with greater semantic significance in general.
\end{itemize}

\subsection{\textbf{\datasetsecond}}

 Prompt for editing scene graphs to create \datasetsecond is shown in Table~\ref{tab:prompt_svg_edit}.

\section{Broader Impacts}
Scene graphs provide useful annotations that can be applied to beyond 2D reasoning, as demonstrated in embodied environments by our experiments. The annotation schema can be broadened to include reasoning over the shapes and structures of 3D objects, enhancing consistency in video understanding, and facilitating image generation with controllable layouts. Accurately creating scene graphs in real-time videos has beneficial implications in fields such as autonomous driving and in helping MLLMs generate culturally sensitive and appropriate remarks by understanding social relationships among people. However, inferring personal relationships and emotional states from scene graphs raises significant privacy and ethical concerns. It is crucial to handle such data responsibly to prevent potential misuse, such as unauthorized surveillance or psychological profiling.
\clearpage

\end{document}